\documentclass[11pt]{mystyle}

\usepackage{amsthm}
\usepackage{xcolor}

\usepackage{mathtools,bm}
\usepackage{algorithm}
\usepackage{algpseudocode}
\usepackage{array,multirow}
\usepackage{longtable} %
\usepackage[numbers,sort&compress]{natbib}
\usepackage[nameinlink,capitalise,noabbrev]{cleveref}
\usepackage{tikz}
\usetikzlibrary{arrows.meta,positioning,fit,calc,shapes.geometric}

\usepackage[export]{adjustbox} %
\usepackage{fontawesome5} %
\usepackage{rotating} %
\usepackage{placeins} %
\usepackage{xspace} %
\usepackage{makecell} %
\usepackage{float} %
\usepackage{needspace} %

\makeatletter
\providecommand{\captionof}[1]{\def\@captype{#1}\caption}
\makeatother

\newcommand{\modellogo}[1]{%
  \IfFileExists{Figure/logo/#1.png}{\includegraphics[width=0.35cm,valign=c]{Figure/logo/#1.png}\,}{}%
}

\definecolor{PipeOrange}{HTML}{EB6834}
\definecolor{PipeBlue}{HTML}{2A78D6}
\definecolor{PipeNavy}{HTML}{1A4F8F}
\definecolor{PipeSlate}{HTML}{64748B}
\definecolor{PipeGrey}{HTML}{E9ECF1}

\definecolor{HeaderBlue}{RGB}{220,235,247}
\definecolor{HeaderAmber}{RGB}{255,242,204}
\definecolor{HeaderRose}{RGB}{252,228,214}
\definecolor{HeaderViolet}{RGB}{233,221,243}
\definecolor{RuleGrey}{RGB}{170,176,188}
\definecolor{OursYellow}{RGB}{255,243,200}

\DeclareRobustCommand{\metricarrow}[2]{%
  \tikz[baseline=-0.05ex,x=1ex,y=1ex]{%
    \draw[black,line width=0.10ex,line cap=round,
      -{Stealth[length=0.50ex,width=0.60ex]}] (0,#1) -- (0,#2);%
  }%
}
\DeclareRobustCommand{\metricup}{\metricarrow{0}{1.3}}
\DeclareRobustCommand{\metricdown}{\metricarrow{1.3}{0}}

\definecolor{TickGreen}{RGB}{40,135,55}
\definecolor{CrossRed}{RGB}{190,46,46}
\definecolor{HalfAmber}{RGB}{210,128,18}
\newcommand{\hmark}{%
  \tikz[baseline=-0.55ex]{%
    \draw[HalfAmber,line width=0.55pt] (0,0) circle (0.55ex);%
    \fill[HalfAmber] (0,0) -- (0,0.55ex) arc[start angle=90,end angle=270,radius=0.55ex] -- cycle;%
  }%
}
\newcommand{\cmark}{\textcolor{TickGreen}{\ding{51}}}
\newcommand{\xmark}{\textcolor{CrossRed}{\ding{55}}}
\newcolumntype{C}{>{\columncolor{HeaderViolet!45}\centering\arraybackslash}m{1.95cm}}
\newcolumntype{B}{>{\columncolor{HeaderBlue!40}\centering\arraybackslash}m{1.63cm}}
\newcolumntype{J}{>{\columncolor{HeaderAmber!55}\centering\arraybackslash}m{1.78cm}}
\newcolumntype{M}{>{\columncolor{HeaderRose!45}\centering\arraybackslash}m{1.85cm}}

\theoremstyle{plain}
\newtheorem{theorem}{Theorem}[section]

\theoremstyle{definition}
\newtheorem{definition}[theorem]{Definition}

\theoremstyle{remark}

\crefname{assumption}{Assumption}{Assumptions}
\Crefname{assumption}{Assumption}{Assumptions}

\newcommand{\benchname}{\textsc{CLE}\xspace}

\usepackage{authblk}

\makeatletter
\apptocmd{\ps@firststyle}{}{}{}
\makeatother

\tcbset{titlebox/.append style={left=4mm,right=4mm}}
\runningtitle{Continual Learning Exam}

\title{Continual Learning Exam\\[-2pt]
  {\fontsize{12}{13}\selectfont Benchmarking Long-Horizon RSI and Experiential Learning in LLM Agents}}

\DeclareRobustCommand{\cleteammark}{%
  \raisebox{-2pt}[0pt][0pt]{\includegraphics[height=11pt]{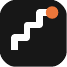}}%
  \hspace{0.4em}%
}
\DeclareRobustCommand{\paperauthorblock}{%
  {\normalfont\normalsize\bfseries \texorpdfstring{\cleteammark}{}CLE Team\par}
}
\author{\paperauthorblock}

\date{}

\hypersetup{hidelinks}

\definecolor{CLELinkAccent}{HTML}{C52273}
\newcommand{\cleresourcerow}[4]{%
  \noindent\makebox[1.55em][c]{#1}\hspace{0.2em}%
  \makebox[8.4em][l]{\sffamily\bfseries #2:}%
  \href{#3}{\textcolor{CLELinkAccent}{\nolinkurl{#4}}}\par
}
\newcommand{\cleprojectlinks}{%
  \par\vspace{5pt}%
  {\normalfont\footnotesize\setlength{\parskip}{1pt}%
    \cleresourcerow{\faGlobe}{Homepage}{https://haoyu-zhao.github.io/CLE/}{https://haoyu-zhao.github.io/CLE/}%
    \cleresourcerow{\includegraphics[height=1.18em,valign=c]{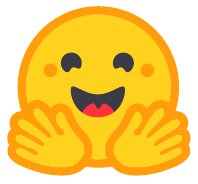}}{Hugging Face}{https://huggingface.co/Continual-Learning-Exam}{https://huggingface.co/Continual-Learning-Exam}%
    \cleresourcerow{\faGithub}{GitHub}{https://github.com/haoyu-zhao/Continual-Learning-Exam}{https://github.com/haoyu-zhao/Continual-Learning-Exam}%
    \noindent\makebox[1.55em][c]{\faEnvelope}\hspace{0.2em}%
    \makebox[8.4em][l]{\sffamily\bfseries Correspondence:}%
    \href{mailto:haoyu.zhao.25@ucl.ac.uk}{\textcolor{CLELinkAccent}{\nolinkurl{haoyu.zhao.25@ucl.ac.uk}}};\ %
    \href{mailto:jun.wang@cs.ucl.ac.uk}{\textcolor{CLELinkAccent}{\nolinkurl{jun.wang@cs.ucl.ac.uk}}}\par
  }%
}
\apptocmd{\abscontent}{\cleprojectlinks}{}{\PackageError{cle}{Could not insert project links}{Check the title-box definition in mystyle.cls.}}
\begin{document}

\begin{abstract}
Continual learning is the ability of LLM agents to improve over time by acquiring, refining, and applying knowledge from experience while retaining useful prior knowledge. It is essential for agents learning over long time horizons, yet current evaluations largely assess individual task performance rather than improvement through accumulated experience. We introduce Continual Learning Exam (CLE), a large-scale benchmark developed by 50+ domain experts to measure experience-driven improvement in realistic professional workflows. \benchname comprises 120 human-authored tasks, each organised as a \emph{five-level ladder}, yielding 600 levels spanning 89 professional roles across 12 occupational domains. Agents must pass each level to unlock the next, with later levels building on experience accumulated during earlier assignments. An evaluation agent assesses each submission against human-authored rubrics. We compare a single-submission setting with one that provides diagnostic feedback after failure and allows one revision. Across 17 harness--model configurations using frontier LLMs, feedback and revision yield a median completion gain of 5.3 percentage points. Yet the best-performing system clears only 47.7\% of levels and completes all five levels in just 23.3\% of tasks, highlighting the difficulty of sustained learning from experience for existing frontier LLMs. As such, the dataset can also be used to post-train LLMs to strengthen their continual learning capabilities.
\end{abstract}
\maketitle

\enlargethispage{48pt}
\begingroup
\setlength{\intextsep}{0pt}
\begin{figure}[H]
  \centering
  \captionsetup{skip=0pt,font=footnotesize}
  \makebox[\linewidth][c]{\includegraphics[width=1.04\linewidth]{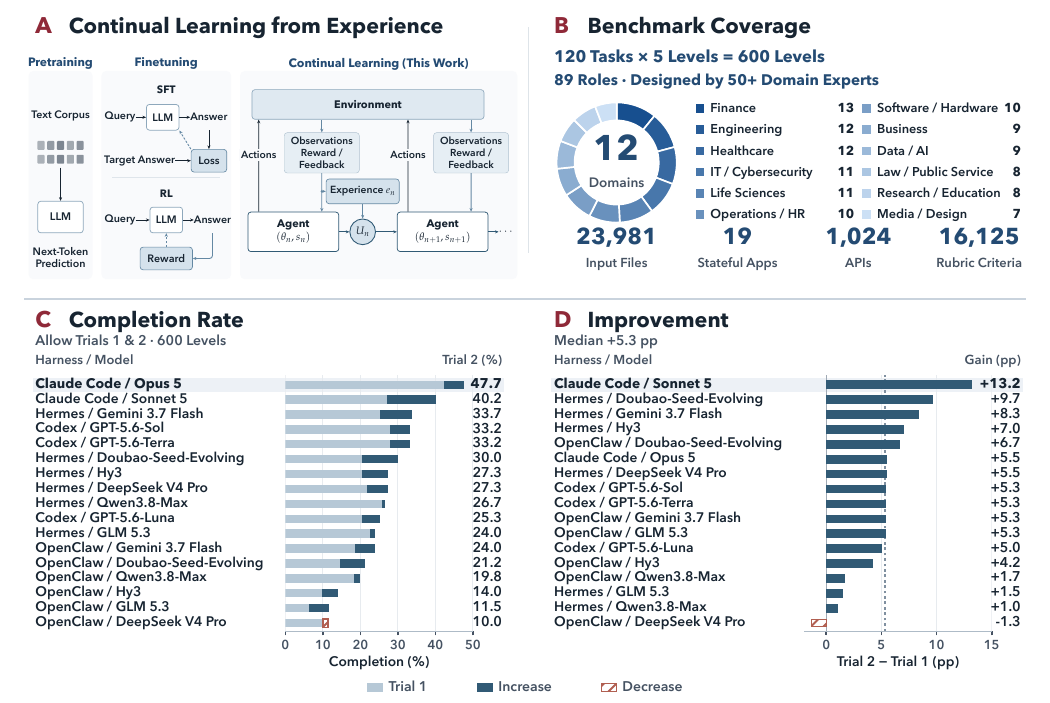}}
  \caption{\textbf{CLE at a Glance.}
  (A) Illustration of continual learning: an agent accumulates experience $e_n$ through interaction and uses the update operator $U_n$ to modify its model parameters $\theta_n$, persistent harness state $s_n$, or both. (B) Summary statistics of CLE. (C) Completion rates across 17 harness--model configurations, measured as the percentage of all 600 levels passed.
  Trial 1 allows one submission per level without diagnostic feedback;
  Trial 2 allows one revision and resubmission following diagnostic feedback on an initial failure.
  (D) The change in completion rate from Trial 1 to Trial 2, in percentage points (pp).}
  \label{fig:front-results}
\end{figure}
\endgroup

\vfill
\begingroup
\normalfont\footnotesize\color{black}
\setlength{\parskip}{1.6pt} %
\hypersetup{hidelinks}
\footnoterule
\noindent\hspace*{1.8em}See the \hyperref[app:authors-affiliations]{\textcolor[HTML]{243B64}{appendix}} for the full author list and affiliations.\par
\endgroup

\clearpage

\section{Introduction}
\label{sec:intro}

For large language model (LLM) agents, \emph{continual learning} unfolds over a long time horizon, as agents acquire, refine, and apply knowledge across successive tasks and interactions while retaining useful prior knowledge~\citep{shi2025continual,parisi2019continual,delange2022continual}. This ability is essential for sustained deployment, enabling agents to adapt to changing environments and task demands while building on past experience and avoiding repeated mistakes. Typically, this accumulated experience can improve agents through updates to model parameters, using supervised fine-tuning or reinforcement learning~\citep{xi2024agentgym,xi2025agentgymrl}, or through adaptations to their harness while keeping model parameters fixed. Harness-based learning includes retaining information in long-term memory~\citep{packer2023memgpt,xu2025amem,chhikara2025mem0}, extracting lessons through reflection~\citep{shinn2023reflexion,zhao2024expel,zhou2025memento}, and developing reusable skills~\citep{wang2023voyager,wang2024awm,tan2024cradle,zhou2026memento}. For example, an agent may learn from a failed tool call, refine its tool-use procedure, and reuse the improved procedure in subsequent tasks. Evaluating continual learning therefore requires tracking performance across a sustained sequence of tasks to determine whether accumulated experience leads to lasting improvements while preserving previously acquired capabilities.

However, most existing agent benchmarks assess individual task performance without directly measuring such improvement. This limitation arises from how they are constructed. General capability benchmarks such as GAIA evaluate whether agents solve individual problems~\citep{mialon2023gaia,jimenez2023swebench,phan2025humanity}. Tasks are presented and scored independently; their scores therefore characterise task-solving performance, rather than how that performance improves through continued learning. Memory benchmarks such as LoCoMo test retrieval and reasoning over interaction histories through question answering~\citep{maharana2024locomo,wu2025longmemeval,wu2026longmemevalv2}, without evaluating how retained knowledge improves subsequent task execution. Benchmarks for Claw-based automation assistants, such as WildClawBench, evaluate agents in executable environments with persistent state, long tool-use trajectories, and cross-application workflows~\citep{ding2026wildclawbench,meng2026clawmark,ye2026claweval}. Persistent state supports continuity in workflow execution, but does not by itself establish that an agent becomes more capable through experience. Evaluating cumulative improvement requires a sequence of task attempts that supports experience reuse under controlled feedback conditions.

To address this evaluation gap, we introduce the Continual Learning Exam (CLE), a large-scale benchmark for evaluating continual learning in realistic human--AI cowork settings. Created by 50+ industry experts, CLE comprises 120 human-authored and human-verified tasks spanning 89 professional roles across twelve professional domains, each organised as a five-level ladder, for a total of 600 levels. These tasks are supported by 23{,}981 real-world input files, 19 simulated applications, and 16{,}125 human-authored evaluation criteria. In the main evaluation setting, agents must pass each level to unlock the next. CLE also supports custom evaluation through blocks, defined as contiguous segments of a task ladder that can begin at any level, with the required support files provided. Later levels introduce new assignments that build on knowledge and experience accumulated through earlier work. We design the task ladders around five forms of knowledge that agents can acquire, retain, and reuse across levels. These five forms reflect common scenarios encountered by LLM agents in real-world workflows. Tools and Skills capture reusable knowledge of how to use tools and carry out tasks, while Data Dependency connects later assignments to earlier information and artefacts. Strategy guides the choice of effective approaches and the avoidance of recurring pitfalls. User Context and Domain Knowledge help align task execution with users' preferences and professional rules and conventions.

 CLE provides a realistic environment for large-scale continual learning evaluation, supported by extensive working materials, stateful software applications, and detailed outcome verification. It includes 23{,}981 real-world input files spanning 77 distinct file extensions. To simulate the software environments used in these assignments, CLE provides 19 stateful mock applications exposing 1{,}024 APIs. Together, these resources support collaborative workflows such as integrating collaborators' changes in Git repositories, coordinating meeting times around participants' availability and preferences, and synchronising project updates across applications. By carrying the effects of earlier actions into later levels of each task, these stateful applications preserve the sequential nature of real-world workflows. Submissions are assessed against human-authored and human-verified rubrics containing 16{,}125 criteria across the benchmark. For each criterion, task authors specify a single, unambiguous expected outcome and the evidence required to verify it. A multi-turn evaluation agent acts as a verifier, using tools to extract evidence from submitted artefacts and final environment state and check it against these predefined outcomes.

We evaluate eleven frontier backbone model variants in 17 configurations across four agent harnesses. All systems execute the same fixed sequence of task ladders with model parameters frozen, retaining reusable harness state across tasks. Trial~1 ($K{=}1$) permits one submission per level without diagnostic feedback. In Trial~2 ($K{=}2$), a failed first submission returns diagnostic feedback based on the rubric, after which the agent can revise and submit once more. Even with diagnostic feedback and an opportunity to revise failed submissions, the best-performing system, Claude Code with Opus~5, completes only 47.67\% of levels. GPT-5.6-Sol and GPT-5.6-Terra each reach 33.17\% under Codex. These results highlight continual learning as a key challenge for \mbox{current agent harnesses}.

Our main contributions are:
\begin{itemize}
  \item We introduce \benchname, a large-scale continual learning benchmark comprising 120 five-level task ladders spanning 89 professional roles across twelve domains, created by 50+ industry experts. The ladders connect distinct assignments through five forms of transferable knowledge, allowing agents to acquire, retain, and reuse experience in realistic human--AI cowork settings.
  \item We develop an executable environment with 23{,}981 real-world input files, 19 stateful mock applications exposing 1{,}024 APIs, and an evaluation agent that grades submissions against 16{,}125 human-authored rubric criteria.
  \item We evaluate 17 configurations spanning 4 agent harnesses and 11 backbone model variants. Diagnostic feedback and an additional submission yield a median completion gain of 5.3 percentage points, yet even the best-performing system completes only 47.67\% of levels.
\end{itemize}

{\color{black}
\hypersetup{allcolors=black}
\ifdefined\headingfont
  \let\cleSavedHeadingFont\headingfont
  \renewcommand{\headingfont}{\cleSavedHeadingFont\color{black}}
\fi
\color{black}
\section{Background and Preliminaries}
\label{sec:background}
\color{black}
\ifdefined\cleSavedHeadingFont
  \let\headingfont\cleSavedHeadingFont
\fi

This section connects the definitions of continual learning with a hierarchy of learning capabilities and a discussion of related concepts and mechanisms.  We first introduce a common formulation that encompasses both updates to model parameters and changes to persistent state, trace the development of continual learning in neural networks and LLM agents, and illustrate how this formulation applies to CLE's task structure (\Cref{subsec:cl-definition}).  Next, we organise continual learning capabilities into five cumulative tiers according to the autonomy and scope of the learning process (\Cref{subsec:cl-hierarchy}). We then discuss concepts related to continual learning in the context of LLM agents (\Cref{subsec:cl-related-concepts}).

\subsection{Continual Learning}
\label{subsec:cl-definition}

Continual learning (CL) aims to improve performance over time by acquiring, refining, and applying knowledge from accumulated experience, while preserving useful prior capabilities~\citep{ring1994continual,thrun1995lifelong,mitchell2015neverending,kirkpatrick2017overcoming,lopezpaz2017gem,mitchell1997machine}.

\Needspace{4\baselineskip}
Let $n=0,1,\ldots$ index learning updates. At learning stage $n$,  a learner with model parameters $\theta_n$ and persistent state $s_n$ uses available experience $e_n$ to update:
\begin{equation}
  (\theta_{n+1},s_{n+1})=U_n(\theta_n,s_n,e_n).
  \label{eq:state-update}
\end{equation}
\begingroup
\setlength{\emergencystretch}{2em}
The initial learner state is $(\theta_0,s_0)$.
Here $s_n$ denotes learning state retained outside the model parameters before the update. In an LLM agent, $s_n$ is the state of the agent's \emph{harness} and can include memories, skills, and other retained information. The harness incorporates relevant parts of this state into the model's context, executes the model's actions, and manages experience.\par
 At stage $n$, $e_n$ denotes the data and signals available for learning that are supplied to $U_n$, whether obtained through interaction or provided externally. These may include training examples or interaction records, such as tool calls and their returned results, together with any available supervision, rewards, or feedback. A single $e_n$ may aggregate multiple interactions and include records retained from earlier updates. The learning update operator $U_n$ updates the learner's state from $(\theta_n,s_n)$ to $(\theta_{n+1},s_{n+1})$ using experience $e_n$, as illustrated in  \Cref{fig:front-results}A. It encompasses parametric learning through changes to model parameters $\theta_n$~\citep{wei2022finetuned,guo2025deepseekr1}, non-parametric learning through changes to persistent state $s_n$~\citep{shinn2023reflexion,wang2023voyager}, or a combination of both.\par
\endgroup

\begingroup
\setlength{\emergencystretch}{2em}
\paragraph{Continual Learning in Neural Networks.}
In 1989, research on sequential neural-network training showed that successive updates to shared weights could overwrite previously learned knowledge, leading to catastrophic interference~\citep{mccloskey1989catastrophic}. Subsequent work on continual and lifelong learning in the 1990s articulated the broader goal of accumulating knowledge and transferring it so that earlier learning facilitates later learning~\citep{ring1994continual,thrun1995lifelong}. To pursue this goal in deep neural networks, representative studies in 2016--2017 explored complementary approaches to preserving earlier capabilities while learning new tasks, including distillation, regularisation, replay, and architectural expansion~\citep{li2016learningwithoutforgetting,kirkpatrick2017overcoming,zenke2017synaptic,lopezpaz2017gem,shin2017generativereplay,rusu2016progressive}. More recent work has explored how knowledge acquired during pretraining can support continual adaptation. Representative approaches keep the backbone fixed and learn compact sets of additional parameters that capture new task knowledge, allowing the system to build on pretrained representations while mitigating forgetting across successive tasks~\citep{wang2022learningtoprompt,smith2023codaprompt}.\par
\endgroup

\begin{definition}[Continual Learning in Neural Networks]
\label{def:neural-cl}
In the classical neural-network setting, continual learning adapts model parameters $\theta_n$ to a sequence of tasks or changing data distributions through $U_n$, using experience $e_n$ and any retained learning state $s_n$, such as stored examples or estimates of parameter importance. It aims to acquire new knowledge while mitigating catastrophic forgetting and supporting knowledge transfer. Catastrophic forgetting refers to the substantial degradation of previously learned capabilities caused by subsequent learning~\citep{lopezpaz2017gem,parisi2019continual}.
\end{definition}

\paragraph{Continual Learning in LLM Agents.}
In 2020, work on large language models demonstrated broad few-shot adaptation through in-context learning, with pretrained models using demonstrations supplied in their context while keeping their parameters fixed~\citep{brown2020fewshot}. An agent can build on this mechanism across interactions by retaining relevant experience and incorporating it into later contexts. From 2023, agent methods extended this route by retaining lessons from feedback and acquiring reusable skills from interaction, allowing earlier experience to guide later attempts and new tasks~\citep{shinn2023reflexion,wang2023voyager}. Subsequent work developed more structured experience management through reusable insights, task procedures, and case memories that can be retrieved and applied across tasks~\citep{zhao2024expel,wang2024awm,zhou2025memento}. In the frozen-model setting, \Cref{eq:state-update} becomes $U_n(\theta_n,s_n,e_n)=(\theta_n,s_{n+1})$: learning changes the persistent harness state that carries knowledge across interactions.

\begin{definition}[Continual Learning in LLM Agents]
\label{def:agent-cl}
Continual learning in LLM agents uses interaction trajectories and feedback to update persistent harness state $s_n$, model parameters $\theta_n$, or both through $U_n$ in \Cref{eq:state-update}. Harness updates retain and refine memories, skills, and other knowledge that the agent reuses in subsequent tasks. These learning updates aim to improve later task performance while preserving useful prior capabilities~\citep{wang2023voyager,zhao2024expel,xi2024agentgym,xi2025agentgymrl}.
\end{definition}

\begin{figure}[!t]
  \centering
  \resizebox{\linewidth}{!}{\definecolor{LadderHeading}{HTML}{294E6A}%
\definecolor{LadderInk}{HTML}{27343B}%
\definecolor{LadderMuted}{HTML}{6B767D}%
\definecolor{LadderRule}{HTML}{DBE1E5}%
\definecolor{LadderSurface}{HTML}{F7F8F9}%
\definecolor{LadderTeal}{HTML}{407B6E}%
\definecolor{LadderTealTint}{HTML}{EDF5F1}%
\definecolor{LadderAmber}{HTML}{A6773A}%
\definecolor{LadderAmberTint}{HTML}{FBF3E6}%
\begin{tikzpicture}[
  x=1pt,y=-1pt,
  font=\normalfont\rmfamily\fontsize{10}{12}\selectfont,
  text=LadderInk,
  label/.style={align=center,inner sep=0pt},
  panel/.style={rounded corners=4pt,draw=LadderRule,fill=LadderSurface,line width=0.8pt},
  stage/.style={rounded corners=3pt,draw=LadderTeal,fill=white,
    line width=0.8pt,minimum width=100pt,minimum height=32pt,inner sep=0pt},
  component/.style={rounded corners=3pt,draw=LadderTeal,fill=white,
    line width=0.85pt,align=center,inner sep=3pt},
  flow/.style={draw=LadderInk,line width=1pt,
    -{Stealth[length=4pt,width=3.3pt]}},
  carry/.style={flow,draw=LadderTeal,line width=1.15pt},
  observe/.style={flow,draw=LadderAmber},
  feedback/.style={flow,draw=LadderAmber},
  record/.style={flow,draw=LadderAmber,dash pattern=on 2.5pt off 1.8pt,line width=0.95pt},
  small/.style={label,font=\normalfont\rmfamily\fontsize{9}{11}\selectfont}
]
\path[use as bounding box] (0,0) rectangle (550,280);

\path[panel] (0,0) rectangle (375,77);
\path[panel] (410,0) rectangle (550,77);
\node[label,text=LadderHeading,font=\normalfont\rmfamily\bfseries\fontsize{11}{13}\selectfont]
  at (187.5,12) {Level $L_i$};
\node[label,text=LadderHeading,font=\normalfont\rmfamily\bfseries\fontsize{11}{13}\selectfont]
  at (480,12) {Level $L_{i+1}$};
\node[small] at (187.5,27) {Current Assignment};
\node[small] at (480,27) {New Assignment};
\foreach \id/\pos/\index in {a/60/n,b/179/{n+1},c/315/m,d/480/{m+1}} {
  \node[stage] (stage\id) at (\pos,54) {};
  \node[label,font=\normalfont\rmfamily\bfseries\fontsize{9.5}{11.5}\selectfont]
    at (\pos,47) {Learning Stage $\index$};
  \node[label,font=\normalfont\rmfamily\fontsize{9.5}{11.5}\selectfont]
    at (\pos,62) {Update $U_{\index}$};
}
\draw[flow] (stagea.east) -- (stageb.west);
\node[label,font=\large] at (247,54) {$\cdots$};
\draw[flow] (stagec.east) -- (staged.west);
\node[small,fill=white,inner sep=1.2pt] at (392.5,42) {Pass};

\draw[carry,rounded corners=3pt] (187.5,77)
  -- (187.5,90) -- (480,90) -- (480,77);
\node[small,fill=white,inner sep=1.5pt] at (330,90)
  {Retain, Reuse, and Refine Knowledge};
\draw[draw=LadderMuted!65,dashed,line width=0.8pt]
  (stagea.south) -- (60,105);

\path[panel,fill=white] (0,105) rectangle (550,279);
\node[label,text=LadderHeading,font=\normalfont\rmfamily\bfseries\fontsize{10.5}{12.5}\selectfont]
  at (275,117) {Within Learning Stage $n$};
\node[component,minimum width=84pt,minimum height=36pt]
  (actor) at (92,179) {\textbf{Actor}\\$(\theta_n,s_n)$};
\node[component,draw=LadderTeal!45,fill=LadderTealTint,
  minimum width=122pt,minimum height=34pt]
  (env) at (300,148)
  {\textbf{Environment}\\{\fontsize{9}{11}\selectfont Tools / Services}};
\node[component,draw=LadderAmber!45,fill=LadderAmberTint,
  minimum width=122pt,minimum height=34pt]
  (evaluator) at (300,210)
  {\textbf{Evaluator}\\{\fontsize{9}{11}\selectfont Rubric-Based Assessment}};

\draw[flow,rounded corners=3pt] (actor.north)
  -- (92,140) -- ([yshift=8pt]env.west);
\node[small,fill=white,inner sep=1.5pt] at (166,140) {Actions};
\draw[observe,rounded corners=3pt] ([yshift=-8pt]env.west)
  -- (149,156) -- (149,171) -- ([yshift=8pt]actor.east);
\node[label,font=\normalfont\rmfamily\fontsize{9}{10.5}\selectfont]
  at (193,171) {Observations\\{\fontsize{8}{9.5}\selectfont Tool Results / State}};
\draw[flow,rounded corners=3pt] ([yshift=-8pt]actor.east)
  -- (149,187) -- (149,202) -- ([yshift=8pt]evaluator.west);
\node[small,fill=white,inner sep=1.5pt] at (195,202) {Submission};
\draw[feedback,rounded corners=3pt] ([yshift=-8pt]evaluator.west)
  -- (92,218) -- (actor.south);
\node[label,font=\normalfont\rmfamily\fontsize{9}{10.5}\selectfont]
  at (166,232)
  {Diagnostic Feedback\\{\fontsize{8}{9.5}\selectfont After First Failure ($K=2$)}};

\node[component,draw=LadderAmber!45,fill=LadderAmberTint,
  minimum width=130pt,minimum height=38pt]
  (experience) at (469,179)
  {\textbf{Experience $e_n$}\\{\fontsize{9}{11}\selectfont Interactions + Feedback}};
\draw[record,rounded corners=3pt] (env.east)
  -- (389,148) -- (389,167) -- ([yshift=12pt]experience.west);
\draw[record,rounded corners=3pt] (evaluator.east)
  -- (389,210) -- (389,191) -- ([yshift=-12pt]experience.west);

\node[circle,draw=LadderTeal,fill=LadderTealTint,line width=0.9pt,
  minimum size=28pt,inner sep=1pt] (update) at (335,254) {$U_n$};
\node[component,minimum width=144pt,minimum height=34pt]
  (updated) at (469,254)
  {{\fontsize{9}{11}\selectfont Next Learning Stage}\\$(\theta_{n+1},s_{n+1})$};
\draw[carry,rounded corners=3pt] (actor.west)
  -- (24,179) -- (24,254) -- (update.west);
\node[small,fill=white,inner sep=1.5pt] at (170,254) {Current State};
\draw[flow,draw=LadderAmber,rounded corners=3pt] (experience.south)
  -- (469,231) -- (335,231) -- (update.north);
\draw[carry] (update.east) -- (updated.west);
\node[small] at (374,244) {Update};
\end{tikzpicture}%
}
  \captionsetup{skip=3pt,font=small}
  \caption{\textbf{Laddered Task Progression in \benchname.} In \benchname, each task comprises five sequential levels. The agent advances to a new assignment only after the current level is passed; retained knowledge is reused and refined across levels. A level $L_i$ can contain multiple learning stages $n,\ldots,m$, followed by stage $m+1$ in $L_{i+1}$. The expanded stage separates two interaction loops: actions produce environment observations, while submissions trigger rubric-based evaluation. After the first failed submission in $K{=}2$, diagnostic feedback returns to the actor to support revision. Dashed arrows collect the resulting interaction history and available feedback as experience $e_n$. Together with the current learner state, $e_n$ informs $U_n$, producing the state for the next learning stage. Stage counts are schematic: $i$ indexes assignments, and $n$ indexes learning updates, not levels or submissions.}
\label{fig:laddered-progression}
\end{figure}
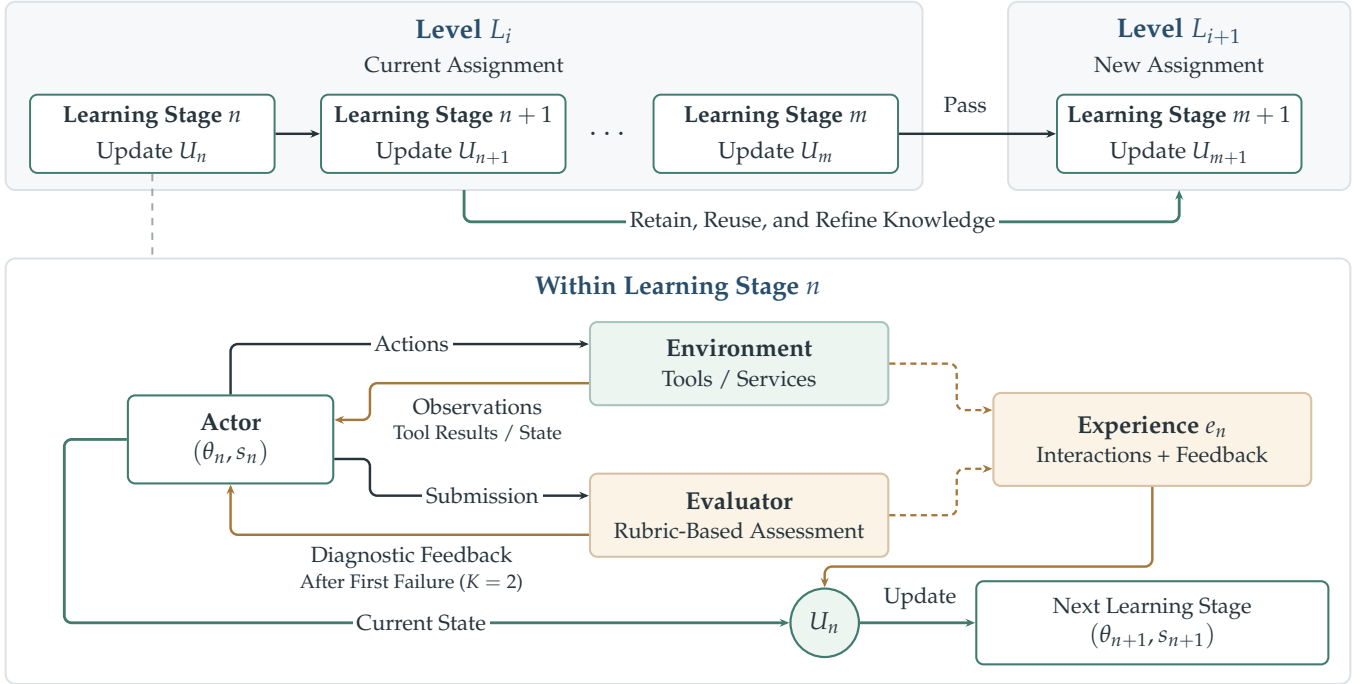

\begingroup
\setlength{\emergencystretch}{2em}
In \benchname, each task comprises five sequential \emph{levels}, $L_1,\ldots,L_5$, representing distinct professional assignments within a shared role and working context. \Cref{fig:laddered-progression} connects this task structure to the general learning formulation in \Cref{eq:state-update}. Here, the level index $i$ identifies assignments defined by \benchname, while $n$ indexes the agent's learning updates. A single level may contain multiple learning updates.\par

Within each level, actions produce environment observations, while submissions trigger rubric-based evaluation. As illustrated in the lower panel, records from one or more interactions and any available feedback provide experience $e_n$. The update operator $U_n$ uses this experience to update model parameters $\theta_n$, persistent harness state $s_n$, or both. Passing the current level unlocks the next assignment, where the agent can reuse and refine knowledge retained through these updates in response to new objectives and requirements. \Cref{sec:bench} provides the full task design.\par
\endgroup

\begingroup
\setlength{\parskip}{2pt}
\setlength{\baselineskip}{12.8pt}
\setlength{\intextsep}{5pt}
\captionsetup{skip=5pt}
\titlespacing*{\subsection}{0pt}{1ex plus1pt minus1pt}{2pt}
\subsection{Capability Hierarchy}
\label{subsec:cl-hierarchy}

We  organise continual learning capabilities into five cumulative \textbf{tiers} (T1--T5), based on  autonomy in learning decisions, adaptation across domains, and autonomy in research design and problem selection (\Cref{fig:cl-hierarchy}). Each tier requires the capabilities of all preceding tiers.

At \textbf{T1}, predefined rules determine which experience the agent learns from, when and how it updates, and how it retains and reuses knowledge. At \textbf{T2}, the agent makes these learning decisions during execution under externally supplied task goals within a given task family, using experience and feedback to check and revise what it has learned. At \textbf{T3}, it adapts this learning process to unfamiliar domains without a new human-designed procedure for each domain, while preserving and transferring useful prior knowledge. At \textbf{T4}, the agent extends these learning capabilities to research by designing investigations to generate and validate new knowledge for an externally supplied research problem, without a supplied investigation or validation plan. It formulates testable subquestions, chooses and carries out experiments, analyses, or proofs, and determines how to check the resulting findings. It retains and reuses verified findings in later investigations. At \textbf{T5}, the agent also sets its own research goals and formulates, selects, and prioritises research problems to pursue, revising its research agenda as verified findings accumulate.

CLE provides a foundation for evaluating agents' continual learning capabilities across T1--T5. In current LLM agent systems such as Claude Cowork and Codex, predefined memory and skill workflows provide the rule-governed foundation described at T1. Within these workflows, agents can choose what experience to retain and reuse, illustrating emerging T2 learning autonomy~\citep{claude2026cowork,codex2026memory,codex2026skills}.

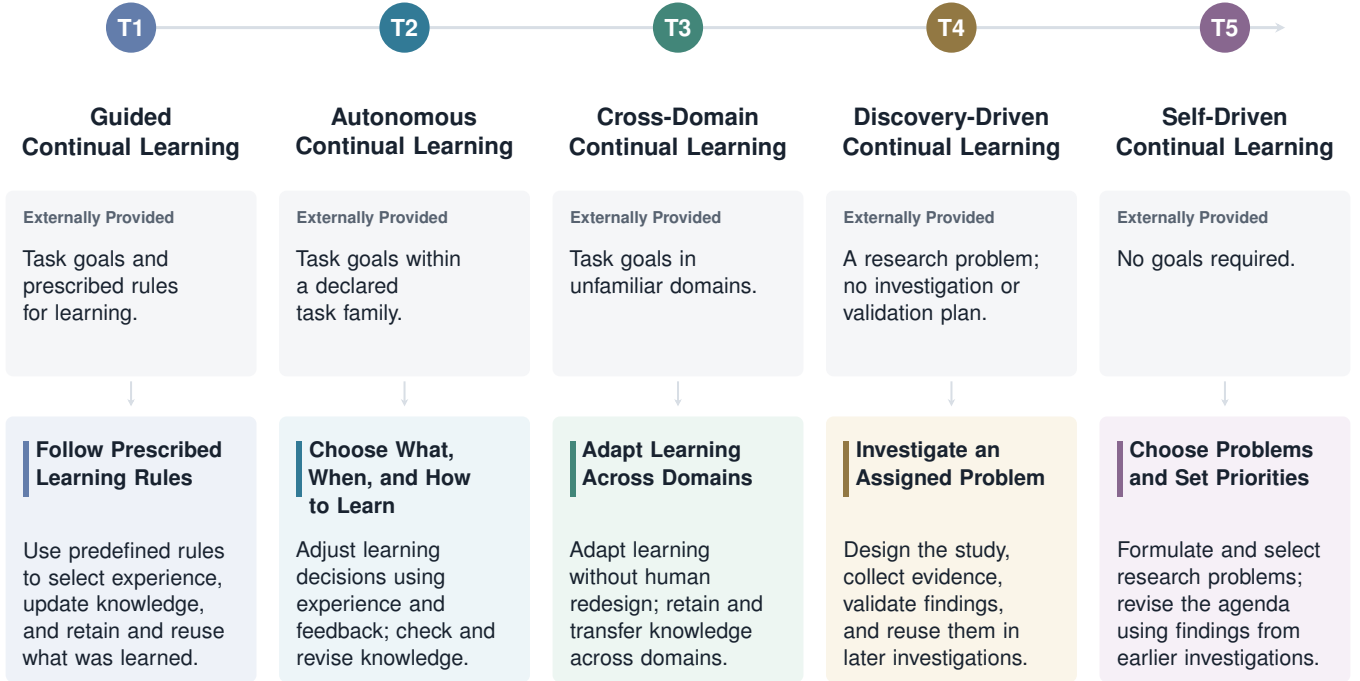
\begin{figure}[H]
  \centering
  \resizebox{\linewidth}{!}{\begin{tikzpicture}[x=1pt,y=-1pt]
  \definecolor{CLInk}{HTML}{18222F}
  \definecolor{CLMuted}{HTML}{576270}
  \definecolor{CLRule}{HTML}{D8DEE5}
  \definecolor{CLSurface}{HTML}{F5F6F8}
  \definecolor{CLAccentOne}{HTML}{637DAB}
  \definecolor{CLAccentTwo}{HTML}{327A96}
  \definecolor{CLAccentThree}{HTML}{428679}
  \definecolor{CLAccentFour}{HTML}{927843}
  \definecolor{CLAccentFive}{HTML}{88678F}
  \definecolor{CLTintOne}{HTML}{EEF2F8}
  \definecolor{CLTintTwo}{HTML}{EDF5F8}
  \definecolor{CLTintThree}{HTML}{EDF6F2}
  \definecolor{CLTintFour}{HTML}{FAF5E9}
  \definecolor{CLTintFive}{HTML}{F6F0F7}
  \path[use as bounding box] (0,0) rectangle (536,273);
  \tikzset{cltext/.style={inner sep=0pt,outer sep=0pt,text=CLInk,
    align=left,anchor=north west,font=\fontfamily{phv}\selectfont\fontsize{8.6}{10.7}\selectfont}}
  \draw[CLRule,line width=1.2pt,-{Stealth[length=4pt,width=4pt]}]
    (50,12) -- (510,12);
  \newcommand{\cleTierCard}[8]{%
    \begin{scope}[shift={(#1,0)}]
      \fill[#7] (50,12) circle (10pt);
      \node[cltext,anchor=center,text=white,
        font=\fontfamily{phv}\selectfont\bfseries\fontsize{9.3}{11.5}\selectfont]
        at (50,12) {#2};
      \node[cltext,anchor=north,align=center,text width=102pt,
        font=\fontfamily{phv}\selectfont\bfseries\fontsize{9.5}{11.8}\selectfont]
        at (50,44) {#3};
      \fill[CLSurface,rounded corners=3pt] (0,77) rectangle (100,151);
      \node[cltext,text=CLMuted,
        font=\fontfamily{phv}\selectfont\bfseries\fontsize{6.5}{8}\selectfont]
        at (7,85) {Externally Provided};
      \node[cltext,text width=87pt] at (7,101) {#4};
      \draw[CLRule,line width=0.8pt,-{Stealth[length=3pt,width=3pt]}]
        (50,153) -- (50,163);
      \fill[#8,rounded corners=3pt] (0,167) rectangle (100,273);
      \fill[#7] (7,177) rectangle (9,199);
      \node[cltext,text width=86pt,
        font=\fontfamily{phv}\selectfont\bfseries\fontsize{8.6}{11.1}\selectfont]
        at (12,177) {#5};
      \node[cltext,text width=87pt] at (7,217) {#6};
    \end{scope}%
  }

  \cleTierCard{0}{T1}
    {Guided\\Continual Learning}
    {Task goals and\\prescribed rules\\for learning.}
    {Follow Prescribed\\Learning Rules}
    {Use predefined rules\\to select experience,\\update knowledge,\\and retain and reuse\\what was learned.}
    {CLAccentOne}{CLTintOne}

  \cleTierCard{109}{T2}
    {Autonomous\\Continual Learning}
    {Task goals within\\a declared\\task family.}
    {Choose What,\\When, and How\\to Learn}
    {Adjust learning\\decisions using\\experience and\\feedback; check and\\revise knowledge.}
    {CLAccentTwo}{CLTintTwo}

  \cleTierCard{218}{T3}
    {Cross-Domain\\Continual Learning}
    {Task goals in\\unfamiliar domains.}
    {Adapt Learning\\Across Domains}
    {Adapt learning\\without human\\redesign; retain and\\transfer knowledge\\across domains.}
    {CLAccentThree}{CLTintThree}

  \cleTierCard{327}{T4}
    {Discovery-Driven\\Continual Learning}
    {A research problem;\\no investigation or\\validation plan.}
    {Investigate an\\Assigned Problem}
    {Design the study,\\collect evidence,\\validate findings,\\and reuse them in\\later investigations.}
    {CLAccentFour}{CLTintFour}

  \cleTierCard{436}{T5}
    {Self-Driven\\Continual Learning}
    {No goals required.}
    {Choose Problems\\and Set Priorities}
    {Formulate and select\\research problems;\\revise the agenda\\using findings from\\earlier investigations.}
    {CLAccentFive}{CLTintFive}
\end{tikzpicture}%
}
  \caption{\textbf{Continual Learning Capability Tiers.} The tiers are cumulative and distinguish learning autonomy, cross-domain adaptation, research design, and problem selection.}
  \label{fig:cl-hierarchy}
\end{figure}

\par
\endgroup

\subsection[Related Concepts and Learning Mechanisms]{Related Concepts and Learning Mechanisms}
\label{subsec:cl-related-concepts}
\begingroup
\setlength{\emergencystretch}{2em}

\par

To clarify how different learning paradigms and mechanisms relate to continual learning, we compare what they retain from experience and how they update model parameters $\theta_n$ or persistent state $s_n$.  \Cref{tab:notions-full} in Appendix~\ref{app:related-concepts} summarises these relationships across learning settings, update and experience-reuse mechanisms, and self-improvement frameworks using the common formulation in \Cref{eq:state-update}.  We write the two output components of $U_n$ as
\[
  \theta_{n+1}=U_n^\theta(\theta_n,s_n,e_n),\qquad
  s_{n+1}=U_n^s(\theta_n,s_n,e_n),
\]
where $U_n^\theta$ and $U_n^s$ return the updated parameters and persistent state, respectively. In the LLM-agent examples, $\theta_n$ denotes backbone parameters, while memories, skills, and auxiliary controller state outside the backbone belong to $s_n$.

\Needspace{5\baselineskip}

Here, we focus on four concepts that clarify complementary aspects of continual learning in LLM agents. Online learning describes the sequential learning setting. Test-time training (TTT) and case-based reasoning (CBR) illustrate optimisation and case reuse, respectively, as mechanisms for incorporating experience. Recursive self-improvement (RSI) concerns improvement of the learning process itself. These perspectives can be combined in the same agent.

\textbf{Online learning} describes a setting in which predictions, feedback, and updates are interleaved as experience arrives~\citep{shalevshwartz2012online}. It does not prescribe a particular implementation of $U_n$: updates may affect $\theta_n$, $s_n$, or both. Within this setting, CL additionally emphasises retaining useful knowledge and applying it to subsequent work.

\textbf{Test-time training (TTT)} specifies an optimisation mechanism: a learning objective constructed from test-time inputs is used to adapt model parameters or adaptive memory~\citep{sun2020testtime,sun2025testtimelayers}. It therefore describes how an update is computed and can operate within an online learning setting. TTT contributes to CL when useful adaptations are retained across subsequent interactions.

\textbf{Case-based reasoning (CBR)} uses explicit past cases: the learner retrieves relevant precedents, adapts their solutions, evaluates the outcome, and retains useful new experience~\citep{aamodt1994cbr}. In LLM agents, the case store and case-selection policy can be updated while keeping the LLM backbone fixed~\citep{zhou2025memento}; these changes belong to $s_n$ in our formulation. CBR specifies an experience-reuse cycle, whereas TTT specifies an optimisation step. The case-selection policy can itself be learned; CBR therefore need not be entirely non-parametric.

\textbf{Recursive self-improvement (RSI)} concerns retained changes that improve how subsequent improvements are generated, evaluated, or implemented~\citep{yudkowsky2007levels,zhao2026memento3}. For example, revised learning or verification procedures in $s_{n+1}$ can change the effective $U_{n+1}$. Online learning, TTT, and CBR can all operate with fixed learning procedures; their use alone does not establish RSI. This distinguishes accumulating task knowledge from improving the process that acquires and refines that knowledge.

\endgroup

}

\FloatBarrier
\section{Continual Learning Exam}
\label{sec:bench}

\begingroup
\emergencystretch=2em
\subsection{Benchmark Design}
\label{subsec:design}
\label{subsec:blocks}

\begin{figure}[tbp]
  \centering
  \includegraphics[width=\linewidth]{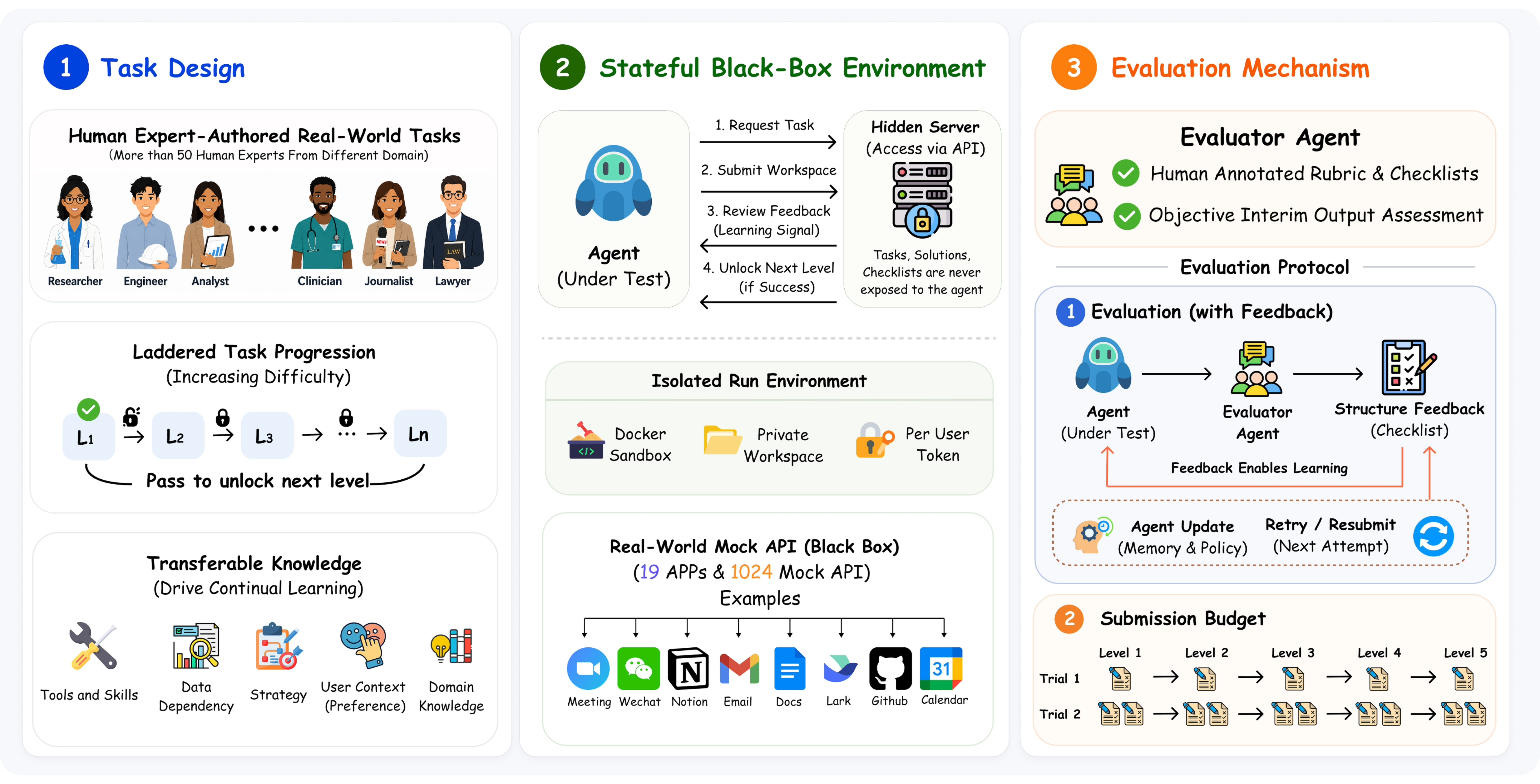}
  \par\vspace{-6pt}
  \caption{\textbf{Overview of \benchname.} Each human-authored task comprises five sequential levels linked by transferable knowledge. Agents interact with stateful applications through APIs, and an evaluation agent verifies submissions against human-authored rubrics. A per-level submission budget of $K{=}1$ allows one submission without diagnostic feedback; $K{=}2$ provides diagnostic feedback after a failed first submission and permits \mbox{one resubmission}.}
  \label{fig:bench-overview}
\end{figure}

\benchname comprises 120 tasks, each representing a realistic human--AI cowork setting and organised as a five-level ladder, yielding 600 human-authored and human-verified levels in total. Developed by domain professionals, each task is grounded in a specific occupational role, and its five levels represent different work assignments that a practitioner in that role might encounter. Each level has a human-authored rubric comprising a structured set of assessment criteria specifying the requirements that a submission must satisfy to pass.  We describe the benchmark design below, using the complete five-level task example in \Cref{fig:task-example} as a concrete illustration.\footnote{More illustrative examples are available on the official CLE website: \url{https://haoyu-zhao.github.io/CLE/}.}

\begin{figure}[!tp]
\centering
  \setlength{\parskip}{0pt}
  \captionsetup{font=small,skip=2pt}
  \includegraphics[width=0.988\textwidth]{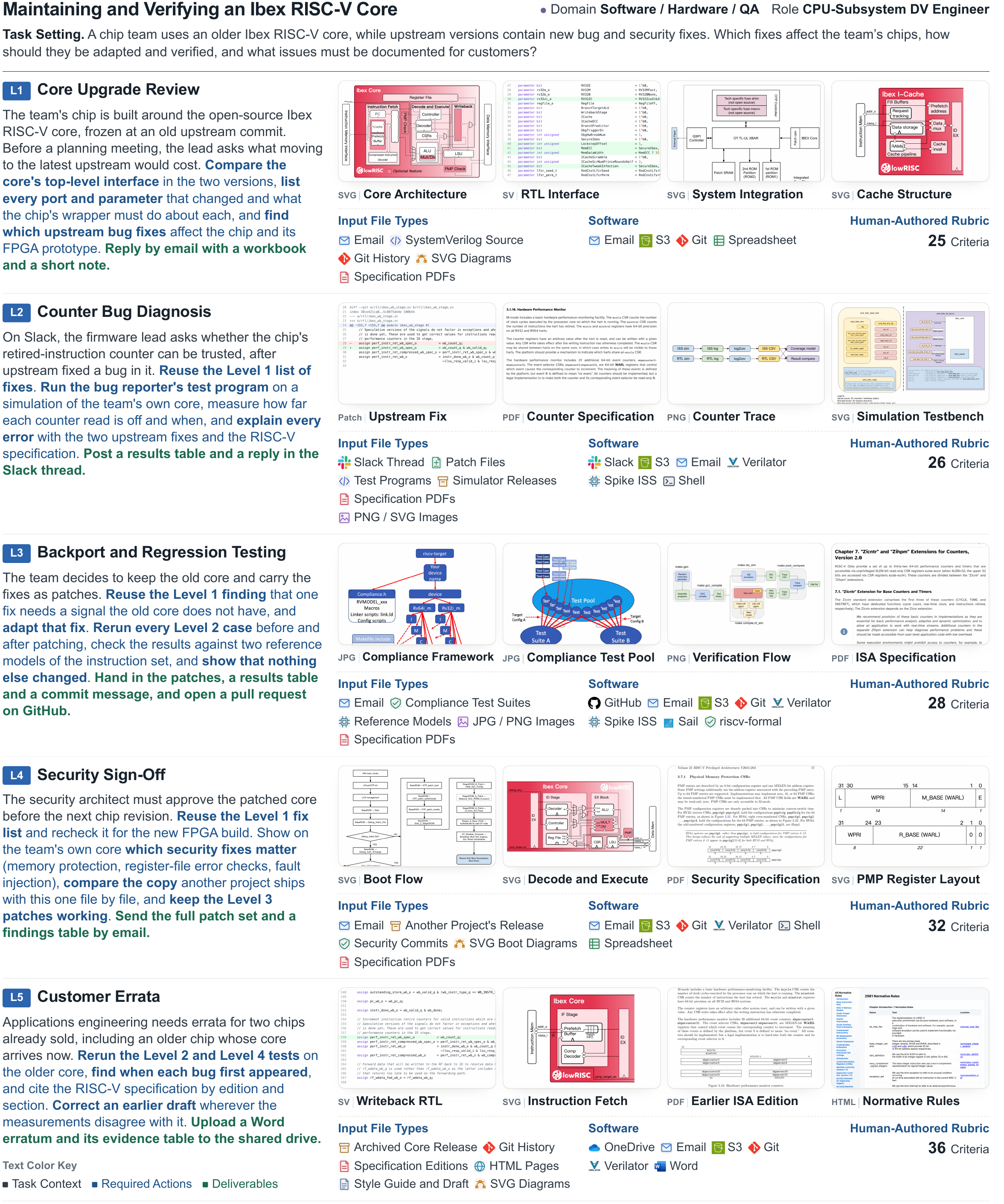}
  \caption{\textbf{A Five-Level Task Example: Ibex RISC-V Core Maintenance and Verification.} The five levels present distinct assignments within a shared professional context, requiring the agent to apply and adapt accumulated knowledge and skills as task objectives and constraints change.}
  \label{fig:task-example}
\par
\end{figure}

\paragraph{Coupled ladders with transferable knowledge.}  We organise each \benchname task as a \textbf{task ladder} with five levels, $L_1,\ldots,L_5$, arranged in sequence like levels in a game (\Cref{fig:laddered-progression}). Each level presents a distinct professional assignment within the same role and working context, and the complete five-level sequence constitutes one benchmark task. Later levels introduce new objectives and expected deliverables, with requirements that can draw on information, constraints, and procedures established in preceding levels. This structure supports evaluating whether agents acquire and retain knowledge from earlier assignments and use it to improve their performance on later ones.  Across the benchmark, we organise these connections around five forms of transferable knowledge. \textbf{Tools and Skills} require agents to apply or adapt tool-use knowledge and procedures acquired during earlier assignments to later work. \textbf{Data Dependency} makes earlier information and artefacts inputs to subsequent deliverables, requiring agents to retrieve and integrate them with current task inputs. \textbf{Strategy} connects earlier experience to later decisions through the reuse of effective approaches and lessons for avoiding recurring pitfalls. \textbf{User Context} requires preferences learned through earlier interactions to inform subsequent decisions and outputs. \textbf{Domain Knowledge} requires professional rules and conventions established or inferred earlier to be applied under new task requirements. These connections extend beyond adjacent levels, requiring agents to identify and use relevant experience from different points in the ladder. Human-authored rubrics specify the outcomes and constraints that submissions must satisfy, allowing the evaluator to assess the resulting deliverables and environment state.
\paragraph{Diagnostic feedback.} Feedback is derived from human-authored rubrics that group detailed checklist items into broader assessment categories. When the submission budget permits, a failed submission receives feedback identifying the relevant rubric category, while individual checklist items and reference answers remain hidden. This feedback provides an explicit learning signal that the actor can use to investigate the problem, revise its approach, and test the revision through another submission. Lessons from this process can be retained in the harness and reused in later levels or tasks to avoid repeating the same mistakes. The grading procedure is described in \Cref{subsec:framework}.

\paragraph{Stateful mock service.} We use stateful mock services to provide agents with a realistic environment for carrying out professional assignments. In everyday work, people rely on applications to access information, update records, and collaborate with others. Our mock services simulate these interactions, allowing agents to perform the corresponding operations through tool calls. Real applications also retain the effects of users' actions as work continues across assignments. We reproduce this continuity by using the same application database throughout a task ladder, so records created or modified in one level remain available in later levels.  \benchname provides \textbf{19} stateful mock applications exposing \textbf{1{,}024} APIs, covering software used by the represented professional roles. These applications support collaborative workflows such as integrating collaborators' changes in shared repositories and coordinating meetings around participants' availability and preferences. Each task supplies real-world input files and specifies the workspace and application state required for its assignments. Each application maintains its own database, initialised separately for each run.

{\color{black}
\hypersetup{allcolors=black}
\paragraph{Block-based evaluation.}
An \textbf{evaluation block} is a contiguous segment of a task ladder. For task $j\in\{1,\ldots,120\}$, a block spanning levels $k$ through $m$ is represented as
\begin{equation}
  B=(j,k,m),\qquad 1\leq k\leq m\leq5.
  \label{eq:block}
\end{equation}
Here $j$ identifies the task, $k$ is the index of the first level, and $m$ is the index of the last level, both inclusive. The block therefore contains $m-k+1$ levels. For example, $B=(7,2,4)$ specifies levels $L_2$, $L_3$, and $L_4$ of task~7. A single-level block has $k=m$, and a full-ladder block has $(k,m)=(1,5)$.

The main experiments use full-ladder blocks $B_j=(j,1,5)$ for all 120 tasks, executed as a single serial stream in a fixed order, with persistent harness state retained across tasks. This state can include memories, skills, and other knowledge retained outside the model parameters. Model parameters remain fixed, while task-specific input files and application databases are initialised for each new task.  Execution starts at level $k$ and advances only after the current level is passed. We evaluate per-level submission budgets of $K{=}1$ and $K{=}2$. If the agent fails to pass within the specified budget, the block ends, its unreached levels through $m$ count as failed, and evaluation proceeds to the next block.

Blocks can also be selected, repeated, or interleaved across tasks.  For a block starting at $k>1$, we manually prepare and load the support files and predecessor artefacts needed to attempt level $k$, so that the agent can start there without executing levels $1$ through $k-1$. Each level retains its original assignment and rubric.
}

\subsection{Agentic Evaluation}
\label{subsec:framework}

\benchname evaluates each submission against a human-authored and human-verified rubric. For each level, the task file describes the assignment, while a separate evaluation file contains the rubric. The evaluation file and its reference materials remain hidden from the agent under test. For each checklist item, task authors specify a concrete target, an expected outcome, and the evidence required for verification. Each item is designed to admit a single, unambiguous pass/fail judgement for a given submission. A multi-turn evaluation agent uses tools to extract evidence from submitted artefacts and authoritative application state and check it against these predefined targets. The evaluator verifies the submitted work against human-verified reference materials rather than independently solving the task to generate its own reference answer. A submission passes only when every checklist item is satisfied. With a per-level submission budget of $K{=}1$, the actor can submit once and receives no diagnostic feedback. With $K{=}2$, a failed first submission returns diagnostic feedback based on the rubric, after which the actor may revise its work and submit once more.

The evaluation agent runs in an isolated grading environment, separate from the agent under test. A unique API credential identifies each run; a unified interface provides task inputs, file transfer, task services, and grading requests, allowing actor execution and grading on separate machines. The actor cannot access the evaluator's private grading materials or workspace. Each grading request starts a fresh Docker container isolated from the actor and other grading executions. Requests are logged to support reproduction, re-grading, and auditing.

\subsection{Problem Formulation}
\label{subsec:actor-evaluator}
\begingroup
\setlength{\emergencystretch}{2em}

The actor is the agent under test, comprising a backbone LLM and a harness that retains experience to guide subsequent actions. It executes assignments and submits work to a separate evaluator, which returns a verdict and any feedback permitted by the protocol. We model the actor's interaction with \benchname as a partially observable Markov decision process (POMDP)~\citep{kaelbling1998planning}. The objective is to maximise the expected number of levels passed over the task stream, subject to execution and submission budgets. \Cref{fig:actor-evaluator-belief} summarises the actor--environment interaction and how the evaluator's verdict determines submission rewards.
\suppressfloats[t]
\begin{figure}[t]
  \centering
  \includegraphics[width=0.96\linewidth]{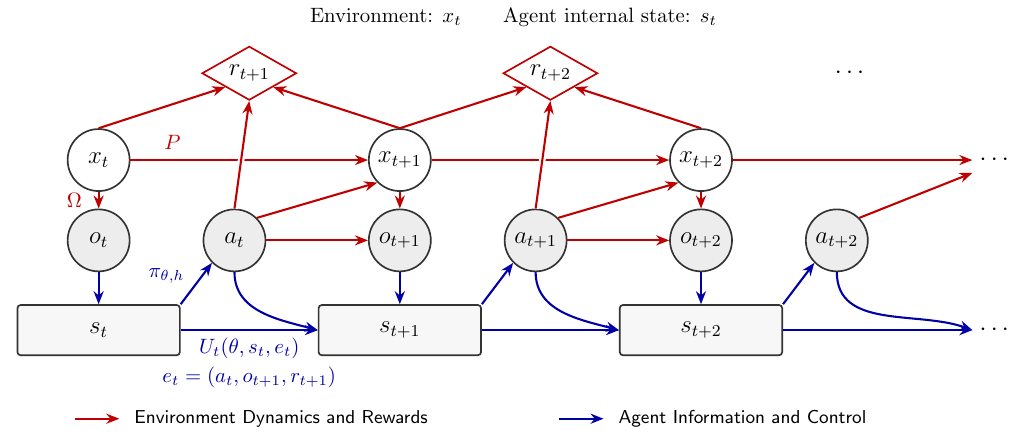}
  \caption{\textbf{Agent--Environment Interaction in CLE.} The policy selects actor action $a_t$ from harness state $s_t$. The environment evolves from $x_t$ to $x_{t+1}$ and returns observation $o_{t+1}$ and reward $r_{t+1}$, which the harness uses with $a_t$ to update $s_t$ to $s_{t+1}$ through $U_t$. Here $e_t=(a_t,o_{t+1},r_{t+1})$ is the interaction experience. On submission, the evaluator's verdict determines $r_{t+1}$. Red arrows show environment dependencies and rewards; blue arrows show agent information and control.}
  \label{fig:actor-evaluator-belief}
\end{figure}

\paragraph{Environment and Observations.}
We consider the stream of full-ladder blocks defined in \Cref{subsec:design}. Each task has five levels, $L_1,\ldots,L_5$, and starts at $L_1$, with $\ell\in\{1,\ldots,5\}$ denoting the current level index.  We index actor actions by $t=0,1,\ldots$. This index does not reset when the actor moves to a new level or task.

 We model the actor--environment interaction as the following POMDP:
\begin{equation}
  \mathcal{M}_A
  =\langle\mathcal{X},\mathcal{A},\mathcal{O},P,\Omega,R\rangle.
  \label{eq:actor-pomdp}
\end{equation}
Here $\mathcal X$, $\mathcal A$, and $\mathcal O$ are the discrete spaces of environment states, actor actions, and observations, respectively.

The actor's external environment comprises the task workspace, stateful applications, and the \benchname evaluation system. The \textbf{environment state} $x_t\in\mathcal X$ includes workspace files and application records, the current task and level, remaining budgets, private grading materials, and the latest tool result or grading outcome.  We assume that $x_t$ contains the information needed to determine the distribution of the next environment state given the actor's action. An \textbf{actor action} $a_t\in\mathcal A$ is an operation on this environment, such as reading a file, updating an application record, or submitting work. An \textbf{observation} $o_t\in\mathcal O$ contains information returned to the actor, such as task instructions, tool results, or permitted grading feedback. Private rubrics and reference materials remain hidden, making execution partially observable.

The \textbf{transition kernel} $P(x'\mid x,a)$ is the probability of the next environment state $x'$ given the current state $x$ and action $a$. The \textbf{observation kernel} $\Omega(o'\mid x',a)$ is the probability of returning observation $o'$ after action $a$ produces state $x'$. Including the latest tool result or grading outcome in $x'$ allows $\Omega$ to return a tool result or grading response together with any newly unlocked assignment. The \textbf{reward function} $R(x,a,x')$ assigns the transition reward, defined in \Cref{eq:actor-reward}.  The benchmark initialises $x_0$ using the first task's prescribed workspace and application state, and provides its assignment as $o_0$.

\paragraph{Policy and Harness Updates.}
The \textbf{harness} constructs the backbone LLM's context, executes actor actions, and manages information retained across interactions.
The \textbf{harness state} $s_t\in\mathcal S$ is the actor's internal state before selecting action $a_t$. It includes the current model context, retained memories and skills, and records of the actor's current plan and progress. The space $\mathcal S$ contains these discrete internal states. For example, an application record belongs to $x_t$, while a summary retrieved into the model context belongs to $s_t$. We take $s_t$ to include all information the harness uses to select its next action and update its state.   The initial state $s_0$ is determined by the harness's initial configuration and $o_0$.

Let $\theta$ denote the backbone parameters and $h$ the harness implementation. Both remain fixed during evaluation, while $s_t$ evolves.  Together, the backbone and harness determine the \textbf{policy} $\pi_{\theta,h}(a\mid s)$, the conditional probability of selecting action $a$ from harness state $s$. The policy and environment generate one interaction as
\begin{equation}
 \begin{aligned}
  a_t&\sim\pi_{\theta,h}(\cdot\mid s_t),\\
  x_{t+1}&\sim P(\cdot\mid x_t,a_t),\\
  o_{t+1}&\sim\Omega(\cdot\mid x_{t+1},a_t),\\
  r_{t+1}&=R(x_t,a_t,x_{t+1}).
 \end{aligned}
 \label{eq:actor-dynamics}
\end{equation}

 Here $r_{t+1}$ denotes the scalar reward for the transition $(x_t,a_t,x_{t+1})$. For non-submission actions, $o_{t+1}$ contains the environment response, such as a tool result, and $r_{t+1}=0$. For submission actions, $o_{t+1}$ contains the evaluator's verdict and any feedback permitted by the protocol; $r_{t+1}$ is 1 for a pass and 0 for a failure (\Cref{eq:actor-reward}). The \textbf{interaction experience} $e_t=(a_t,o_{t+1},r_{t+1})$ records the action, observation, and reward, providing the data and signals available for learning from this interaction. The harness uses $s_t$ and $e_t$ to update:
 \begin{equation}
  (\theta,s_{t+1})=U_t(\theta,s_t,e_t).
  \label{eq:harness-update}
\end{equation}
The harness $h$ implements the possibly stochastic \textbf{update operator} $U_t$. Internal harness operations, such as retrieving a memory or editing a retained skill, are included in action selection and state updates. The policy $\pi_{\theta,h}$ uses the resulting $s_{t+1}$ to select the next action.

After each interaction, $U_t$ maintains the actor's context and progress records and may retain or refine knowledge for later use, such as a discovered constraint or a procedure for checking outputs. The index $t$ counts actor actions, while $n$ in \Cref{eq:state-update} counts learning updates. One learning update can draw on several interactions, and one task level can contain several learning updates. Retained knowledge affects subsequent action probabilities through $\pi_{\theta,h}(\cdot\mid s_t)$.

\paragraph{Submission Evaluation, Rewards, and Progression.}
A submission action $a_t$ consumes one attempt and invokes the evaluator described in \Cref{subsec:framework}. The actor waits while the evaluator checks the submitted artefacts and an application-state snapshot against the current level's private rubric and reference materials. These inputs remain fixed during verification. Each submission uses a fresh grading workspace under the same evaluator configuration. The evaluator returns a verdict $v_t\in\{0,1\}$ and a grading report. A submission passes ($v_t=1$) only when every rubric item is judged satisfied.

The environment records the verdict and report in $x_{t+1}$. For submission actions, $P$ therefore includes the distribution of grading outcomes and their effects on task progression.  The observation $o_{t+1}$ contains the verdict and feedback permitted by the protocol. The reward is
\begin{equation}
 r_{t+1}=R(x_t,a_t,x_{t+1})=
 \begin{cases}
  v_t, & \text{if }a_t\text{ submits work for grading},\\
  0, & \text{otherwise}.
 \end{cases}
 \label{eq:actor-reward}
\end{equation}
Given a transition $(x_t,a_t,x_{t+1})$, the reward is determined by the stored verdict for submission actions and is zero otherwise. A passing submission receives reward 1; all other actions receive 0.  A failed submission has reward 0 but can still provide diagnostic feedback identifying the relevant rubric category, as specified by the protocol in \Cref{subsec:framework}. This feedback is part of $o_{t+1}$ and hence $e_t$, allowing the harness to use it for revision and learning.

Each level allows at most $K$ submissions, with $K\in\{1,2\}$ in the main evaluation settings. A pass at $L_\ell$ unlocks $L_{\ell+1}$ when $\ell<5$; after a failure, the actor can revise at the same level if attempts remain. On entering a new level, the environment updates the assignment and resets its submission budget to $K$, while carrying forward existing task artefacts and application records. Passing $L_5$ or failing with no attempts remaining ends the task. Unreached levels count as failed, and execution proceeds to $L_1$ of the next task, if any. When the next task starts, the environment initialises its files and application databases, while the harness carries retained knowledge into the new task and manages its current context. Execution budgets also constrain the duration of the task stream.

\paragraph{Joint Dynamics and Evaluation Objective.}
With $\theta$ and $h$ fixed, the \textbf{joint environment--actor state} $(x_t,s_t)$ evolves as a Markov process under the rules above. For a nonterminal current state $(x,s)$, let $K_{A,t}(x',s'\mid x,s)$ denote the conditional probability of the next joint state $(x',s')$. For each action $a$ and returned observation $o'$, \Cref{eq:actor-dynamics,eq:harness-update} give the product of action selection, environment transition, observation, and harness-update probabilities. Summing over $a$ and $o'$ gives
 \begin{equation}
\begin{aligned}
 K_{A,t}(x',s'\mid x,s)
 ={}&\sum_{\substack{a\in\mathcal A\\o'\in\mathcal O}}
 \pi_{\theta,h}(a\mid s)\,P(x'\mid x,a)\,\Omega(o'\mid x',a)\\
 &\quad\times \Pr\!\left[U_t\bigl(\theta,s,(a,o',R(x,a,x'))\bigr)=(\theta,s')\right].
\end{aligned}
\label{eq:actor-kernel}
\end{equation}
Here $x,x'\in\mathcal X$ and $s,s'\in\mathcal S$.  The last factor is the probability that the harness produces $s'$ from current state $s$ and experience $(a,o',R(x,a,x'))$, with randomness specified by $U_t$. This randomness can arise from sampling model outputs. If the update retains outputs generated during action selection, this factor uses their conditional distribution given $s$ and the selected action $a$. Each factor is a normalised conditional distribution, so $\sum_{x',s'}K_{A,t}(x',s'\mid x,s)=1$. The index $t$ allows the harness update to depend on the interaction step.  This kernel includes submission evaluation through $P$, feedback through $\Omega$, and their use in the harness update through $e_t$.

The sampling and update rules in \Cref{eq:actor-dynamics,eq:harness-update}, together with initialisation and the reward rule, specify how an evaluation run unfolds. Let $T$ be the number of actor steps before the task stream terminates under the protocol and execution budgets. The actor's objective is the expected undiscounted cumulative reward,
\begin{equation}
 J(\theta,h)=\mathbb{E}_{\theta,h}\!\left[\sum_{t=0}^{T-1}r_{t+1}\right].
 \label{eq:actor-objective}
\end{equation}
The expectation covers  action selection, environment responses, and harness updates. Each passed level contributes reward 1 exactly once. The total reward in a run therefore equals the number of levels passed. The reported completion rate is the number of levels passed divided by 600. Holding $\mathcal M_A$ and the backbone fixed, different harnesses change $\pi_{\theta,h}$ and $U_t$, thereby changing the induced joint dynamics and expected return. Within a run, experience retained in $s_t$ influences later actions and their rewards.

\endgroup

\subsection{Data Construction}
\label{subsec:task-construction}

The task environments and grading specifications are authored by domain professionals. We recruited \textbf{50+} industry experts across a broad range of knowledge-intensive occupations. They include PhD researchers across disciplines, software and hardware engineers, accountants, and financial and operations analysts, among others. Each expert authors benchmark tasks around work from their own role that would benefit from agent assistance and grounds those tasks in authentic input files rather than AI-generated artefacts. The data construction pipeline is summarised in \Cref{fig:construction}.

\begin{figure}[htbp]
  \centering
  \begin{tikzpicture}[
    box/.style={rounded corners=2.5pt, draw=RuleGrey, line width=0.5pt, fill=#1,
                align=center, font=\small, inner sep=4pt,
                text width=21mm, minimum height=13.5mm},
    box/.default=white,
    flow/.style={-{Stealth[length=4.5pt,width=3.5pt]}, draw=RuleGrey, line width=0.7pt},
    back/.style={-{Stealth[length=4.5pt,width=3.5pt]}, draw=CrossRed, line width=0.7pt,
                 dash pattern=on 2.5pt off 1.8pt},
    merge/.style={draw=CrossRed, line width=0.7pt, dash pattern=on 2.5pt off 1.8pt},
    lbl/.style={font=\scriptsize, text=CrossRed, fill=white, inner xsep=1.5pt, inner ysep=0.5pt}
  ]
  \node[box=HeaderViolet!55] (design) {Expert\\Design};
  \node[box=HeaderBlue!55, right=5mm of design] (submit) {Submit to\\Review};
  \node[box=HeaderAmber!70, right=5mm of submit] (review) {Agent Auto\\Review \&\\Real Run};
  \node[box=HeaderRose!55, right=5mm of review] (human) {Dual Human\\Review};
  \node[box=OursYellow, right=5mm of human] (admit) {Admitted\\to \benchname};

  \draw[flow] (design) -- (submit);
  \draw[flow] (submit) -- (review);
  \draw[flow] (review) -- (human);
  \draw[flow] (human) -- (admit);

  \coordinate (trunk) at ($(human.south)+(0,-6.5mm)$);
  \draw[back] (human.south) -- ++(0,-6.5mm) -| (design.south);
  \draw[merge] (review.south) -- (review.south |- trunk);
  \fill[CrossRed] (review.south |- trunk) circle (0.7pt);
  \node[lbl, anchor=south] at ($(design.south|-trunk)!0.5!(review.south|-trunk)+(0,0.6mm)$)
    {Feedback};
  \end{tikzpicture}
  \caption{\textbf{Data Construction Pipeline.} A domain expert drafts each task from a workflow in their own professional practice. The draft is audited by the review agent and cross-reviewed by two randomly assigned human reviewers, and it enters \benchname only after clearing both gates. A rejection at either gate returns the task to its author for revision.}
  \label{fig:construction}
\end{figure}
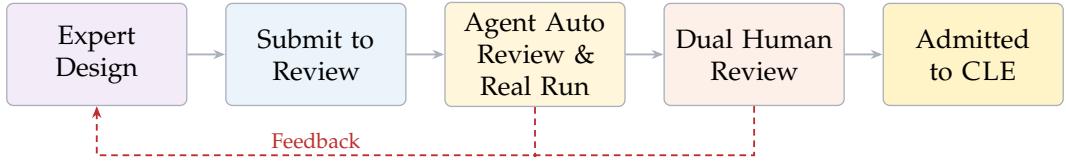

For each level, the task designer defines the occupational role and working context, provides the input artefacts, specifies the expected deliverable, and writes a checklist of criteria that a passing submission must satisfy. A task-review agent then audits the draft using a review skill authored and refined by human experts over multiple iterations. Following the skill's predefined review criteria, the agent checks whether each task meets the benchmark's design requirements. The review includes checks on format compliance, the authenticity of input files, answer leakage, rubric verifiability, and continual-learning design. The review agent also switches to an actor role and runs the task end to end through the same interface available to an evaluated agent. This real execution verifies that the task functions as intended, poses a realistic and non-trivial challenge, and exhibits a clear progression in difficulty across levels. These checks help keep the benchmark challenging for current agents, leaving room to measure gains from continued learning. Each review returns one of five conference-style verdicts, Strong Accept, Accept, Borderline, Major Revision, or Reject, together with itemised findings and concrete revision suggestions. Task designers revise and resubmit their drafts until they meet the acceptance standard. We provide further details in Appendix~\ref{app:auto-review}.

Tasks receiving Accept or higher proceed to final human review. Each task is assigned two randomly selected reviewers, who independently conduct a full manual review. A task is admitted only if both reviewers accept it; otherwise, it returns to the task designer for revision. The 120 tasks that clear both the automated and human review stages constitute the released benchmark.

\subsection{Data Overview}
\label{subsec:task-overview}

\benchname contains \textbf{120} tasks spanning occupational roles from twelve domains, with between 7 and 13 tasks in each domain (\Cref{fig:composition}A). Each task is organised as a ladder of \textbf{five} gated levels, for \textbf{600} levels in total. Tasks are written in the working language of their contributors, with 508 levels in English and 92 in Chinese.

The released benchmark contains 23{,}981 files, including spreadsheets, JSON records, PDF documents, email messages, images, audio recordings, and source code (\Cref{fig:composition}B). Task inputs consist of real-world files spanning 77 extensions and retain noise such as duplicate records and conflicting information. Submissions are evaluated against \textbf{16{,}125} human-authored checklist items, averaging 26.9 per level and 134.4 per task. The number of checklist items increases across levels (\Cref{fig:composition}C).  \Cref{fig:task-example} illustrates these inputs and rubric counts for a complete five-level task.

\begin{figure}[tbp]
  \centering
  \includegraphics[width=\linewidth]{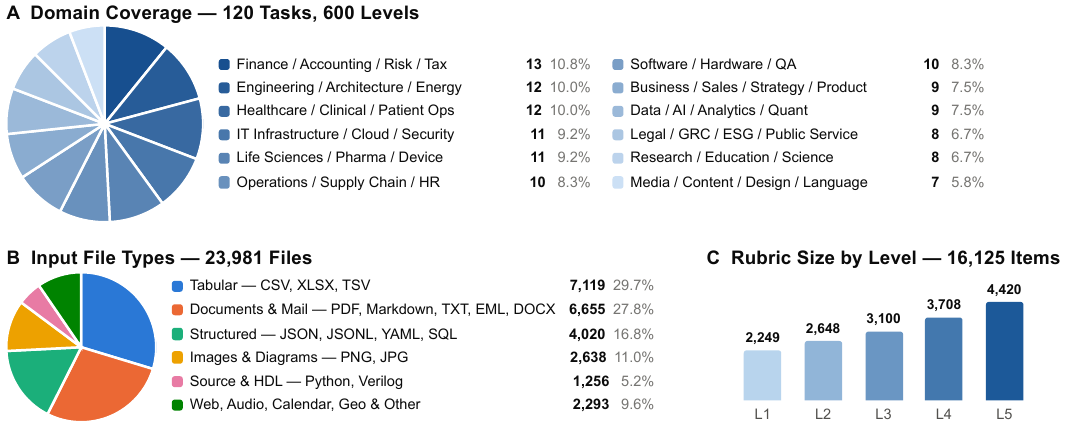}
  \caption{\textbf{\benchname Statistics.} \textbf{(A)} The 120 tasks (600 levels) grouped into twelve occupational families and counted in tasks. \textbf{(B)} The 23{,}981 input files grouped by format family, counted in files. \textbf{(C)} The 16{,}125 human-authored checklist items by level, counted in items summed over the 120 tasks.}
  \label{fig:composition}
\end{figure}

\par\endgroup
\section{Experiment}
\label{sec:exp}

\subsection{Evaluation Setup}
\label{subsec:setup}

\begingroup
\emergencystretch=3em
\paragraph{Systems under evaluation.} We evaluate \textbf{17} agent configurations spanning four agent harnesses and eleven backbone models (\Cref{tab:main-results}). The harnesses are Claude Code~\citep{claudecode}, Codex~\citep{codex}, Hermes-Agent~\citep{hermesagent}, and OpenClaw~\citep{openclaw}. The backbone models are GPT-5.6-Sol~\citep{gpt56sol}, GPT-5.6-Terra, GPT-5.6-Luna, Claude Opus 5~\citep{claudeopus5}, Claude Sonnet 5~\citep{claudesonnet5}, Gemini 3.7 Flash~\citep{gemini37flash}, \mbox{Doubao-Seed-Evolving}~\citep{doubaoseedevolving}, Qwen3.8-Max~\citep{qwen38max}, Hy3~\citep{hunyuan3}, DeepSeek V4 Pro~\citep{deepseekv4}, and GLM 5.3~\citep{glm53}.
\par\endgroup

\paragraph{Evaluation protocol.} Every system is run over the full benchmark as a single serial stream of all 120 ladders in one fixed order. The agent under test uses the benchmark API provided by \benchname (\Cref{subsec:framework}) as its sole interface for retrieving tasks and input files, observing the task environment, and submitting deliverables for grading. In Trial~1 ($K{=}1$), each agent can submit once per level without diagnostic feedback. In Trial~2 ($K{=}2$), it can use diagnostic feedback from a failed first submission to revise and resubmit once. The trial labels denote two complete benchmark runs with different per-level submission budgets.

Claude Code and Codex use official Anthropic and OpenAI subscriptions, respectively, while the remaining agents access their models through a unified OpenRouter endpoint. For reasoning effort and thinking mode, we use the default settings of the respective agents and models. Every verdict is produced by the evaluation agent of \Cref{subsec:framework}, instantiated over DeepSeek~V4~Flash~\citep{deepseekflash}. To assess alignment with human grading, we compare the evaluator's judgements with adjudicated human labels on a validation sample. The evaluator achieves 96.86\% agreement at the rubric-item level (Appendix~\ref{app:human-agent-agreement}, \Cref{tab:human-agent-agreement}). The task design supports this agreement. Each checklist item specifies an unambiguous expected outcome and the evidence required for verification.

\paragraph{Metrics.} We evaluate every system on all 120 tasks, and a level the agent never unlocks is scored as failed. The per-level pass rate $L_k$ is the proportion of tasks whose first $k$ levels are all cleared within the trial's budget, so $L_5$ is the proportion of tasks completed in full. Overall completion (Compl.\,\%) is the proportion of the 600 levels cleared. In addition to these pass rates, we report the resources each trial consumes along three axes. \textbf{Actor Tok.} reports the total number of tokens consumed by the agent under test. \textbf{Time} is the cumulative agent execution time, obtained by summing per-task active runtime over the 120 tasks; we sum active runtime rather than the wall-clock span of the run so that queueing delays, provider throttling, and parallelism across machines do not affect the measurement. \textbf{Cost} is the corresponding end-to-end expenditure in USD.
\subsection{Results and Analysis}
\label{subsec:results}

\Cref{tab:main-results} reports the results for all 17 agent configurations in Trials~1 and~2. Completing all five levels of a task remains challenging for current agents, even with diagnostic feedback and an opportunity to revise their work. We provide a more detailed analysis in Appendix~\ref{app:additional-results}.

Comparing Trials~1 and~2 assesses diagnostic feedback and an additional revision opportunity together. A separate matched comparison that resets the harness to its initial knowledge state before each task, while holding the model, tasks, feedback policy, and execution budgets fixed, would assess the contribution of knowledge retained across tasks.

\begin{table}[t]
\hypersetup{allcolors=black}\color{black}
\centering
\caption{\textbf{Main results on \benchname.} Rows are grouped by agent harness, with one backbone model per row. Trial~1 allows one submission per level. Trial~2 allows a second submission after diagnostic feedback on a failed first submission. Completion and level pass rates are reported as percentages. \metricup/\metricdown{} indicate whether higher/lower is preferred.}
\label{tab:main-results}
\renewcommand{\arraystretch}{1.3}
\setlength{\tabcolsep}{3.5pt}
\resizebox{\textwidth}{!}{%
\begin{tabular}{@{} l *{18}{c} @{}}
\toprule
\addlinespace[2pt]
\multirow{3}{*}{\textbf{Setting}} & \multicolumn{9}{c}{\textbf{Trial 1}} & \multicolumn{9}{c}{\textbf{Trial 2}} \\
\cmidrule(l{2pt}r{2pt}){2-10}\cmidrule(l{2pt}r{2pt}){11-19}
 & \multirow{2}{*}{\textbf{Compl.\hspace{0.25em}\metricup}} & \multicolumn{5}{c}{\textbf{Per-Level Pass\hspace{0.25em}\metricup}} & \multirow{2}{*}{\textbf{\makecell{Actor\\Tok.\hspace{0.25em}\metricdown}}} & \multirow{2}{*}{\textbf{Time\hspace{0.25em}\metricdown}} & \multirow{2}{*}{\textbf{Cost\hspace{0.25em}\metricdown}} & \multirow{2}{*}{\textbf{Compl.\hspace{0.25em}\metricup}} & \multicolumn{5}{c}{\textbf{Per-Level Pass\hspace{0.25em}\metricup}} & \multirow{2}{*}{\textbf{\makecell{Actor\\Tok.\hspace{0.25em}\metricdown}}} & \multirow{2}{*}{\textbf{Time\hspace{0.25em}\metricdown}} & \multirow{2}{*}{\textbf{Cost\hspace{0.25em}\metricdown}} \\
\cmidrule(l{2pt}r{2pt}){3-7}\cmidrule(l{2pt}r{2pt}){12-16}
 &  & \textbf{L1} & \textbf{L2} & \textbf{L3} & \textbf{L4} & \textbf{L5} &  &  &  &  & \textbf{L1} & \textbf{L2} & \textbf{L3} & \textbf{L4} & \textbf{L5} &  &  &  \\
\midrule

\multicolumn{19}{c}{\cellcolor{HeaderViolet}\textbf{Claude Code}}\\
\addlinespace[1pt]
\modellogo{anthropic}Claude Opus 5               & 42.17\% & 79.17\% & 49.17\% & 37.50\% & 25.00\% & 20.00\% & 380.87M & 32.80h & \$410.92 & 47.67\% & 85.83\% & 56.67\% & 39.17\% & 33.33\% & 23.33\% & 520.62M & 58.90h & \$720.56 \\
\modellogo{anthropic}Claude Sonnet 5             & 27.00\% & 65.00\% & 35.00\% & 20.00\% & 8.33\% & 6.67\% & 470.51M & 19.25h & \$220.95 & 40.17\% & 74.17\% & 48.33\% & 34.17\% & 25.00\% & 19.17\% & 1008.60M & 37.10h & \$506.30 \\
\addlinespace[3pt]

\multicolumn{19}{c}{\cellcolor{HeaderViolet}\textbf{Codex}}\\
\addlinespace[1pt]
\modellogo{openai}GPT-5.6-Sol                 & 27.83\% & 58.33\% & 32.50\% & 23.33\% & 13.33\% & 11.67\% & 131.30M & 23.10h & \$168.32 & 33.17\% & 65.00\% & 38.33\% & 26.67\% & 18.33\% & 17.50\% & 152.30M & 28.20h & \$210.83 \\
\modellogo{openai}GPT-5.6-Terra               & 27.83\% & 62.50\% & 31.67\% & 20.00\% & 13.33\% & 11.67\% & 130.34M & 18.60h & \$73.49 & 33.17\% & 68.33\% & 40.83\% & 25.83\% & 18.33\% & 12.50\% & 135.20M & 20.20h & \$83.16 \\
\modellogo{openai}GPT-5.6-Luna                & 20.33\% & 50.83\% & 25.83\% & 13.33\% & 7.50\% & 4.17\% & 130.60M & 16.20h & \$28.62 & 25.33\% & 56.67\% & 30.83\% & 19.17\% & 11.67\% & 8.33\% & 133.50M & 19.10h & \$33.53 \\
\addlinespace[3pt]

\multicolumn{19}{c}{\cellcolor{HeaderViolet}\textbf{Hermes-Agent}}\\
\addlinespace[1pt]

\modellogo{gemini}Gemini 3.7 Flash
& 25.33\% & 55.00\% & 33.33\% & 19.17\% & 10.83\% & 8.33\%
& 486.25M & 46.10h & \$215.64
& 33.67\% & 62.50\% & 40.83\% & 30.00\% & 19.17\% & 15.83\%
& 901.82M & 102.20h & \$394.84 \\
\modellogo{doubao}\textcolor{black}{Doubao-Seed-Evolving} &
20.33\% & 48.33\% & 27.50\% & 14.17\% & 6.67\% & 5.00\%
& 232.01M & 64.40h & \$74.05
& 30.00\% & 62.50\% & 38.33\% & 22.50\% & 14.17\% & 12.50\% & 345.81M & 81.20h & \$102.61 \\

\modellogo{qwen}Qwen3.8-Max
& 25.67\% & 70.83\% & 33.33\% & 15.83\% & 5.83\% & 2.50\%
& 267.82M & 65.17h & \$118.71
& 26.67\% & 74.17\% & 34.17\% & 16.67\% & 5.83\% & 2.50\% & 286.43M & 70.12h & \$134.54 \\

\modellogo{tencent}Hy3
& 20.33\% & 50.83\% & 26.67\% & 13.33\% & 6.67\% & 4.17\%
& 193.69M & 35.90h & \$8.26
& 27.33\% & 60.83\% & 35.00\% & 19.17\% & 11.67\% & 10.00\%
& 390.92M & 80.30h & \$16.18 \\

\modellogo{deepseek}DeepSeek V4 Pro
& 21.83\% & 56.67\% & 27.50\% & 15.00\% & 5.83\% & 4.17\%
& 312.56M & 25.70h & \$44.81
& 27.33\% & 55.83\% & 33.33\% & 20.00\% & 14.17\% & 13.33\%
& 685.93M & 85.00h & \$96.51 \\

\modellogo{glm}GLM 5.3
& 22.50\% & 60.00\% & 27.50\% & 13.33\% & 7.50\% & 4.17\%
& 290.76M & 41.50h & \$82.61
& 24.00\% & 62.50\% & 30.00\% & 15.00\% & 7.50\% & 5.00\%
& 320.54M & 48.90h & \$102.58 \\

\addlinespace[3pt]

\multicolumn{19}{c}{\cellcolor{HeaderViolet}\textbf{OpenClaw}}\\
\addlinespace[1pt]

\modellogo{gemini}Gemini 3.7 Flash            & 18.67\% & 48.33\% & 24.17\% & 12.50\% & 5.83\% & 2.50\% & 323.78M & 16.30h & \$182.65 & 24.00\% & 59.17\% & 32.50\% & 15.83\% & 6.67\% & 5.83\% & 524.45M & 23.77h & \$212.33 \\
\modellogo{doubao}Doubao-Seed-Evolving
& 14.50\% & 40.00\% & 15.00\% & 8.33\% & 5.00\% & 4.17\% & 150.83M & 33.10h & \$32.60
& 21.17\% & 48.33\% & 26.67\% & 15.83\% & 8.33\% & 6.67\% & 172.62M & 31.30h & \$40.50 \\
\modellogo{qwen}Qwen3.8-Max & 18.17\% & 60.00\% & 18.33\% & 9.17\% & 1.67\% & 1.67\% & 227.52M & 52.42h & \$92.51 & 19.83\% & 65.00\% & 20.00\% & 10.83\% & 1.67\% & 1.67\% & 245.38M & 58.42h & \$101.54 \\
\modellogo{tencent}Hy3                         & 9.83\% & 31.67\% & 7.50\% & 4.17\% & 3.33\% & 2.50\% & 190.89M & 20.40h & \$10.27 & 14.00\% & 37.50\% & 12.50\% & 7.50\% & 6.67\% & 5.83\% & 198.80M & 26.10h & \$11.95 \\
\modellogo{deepseek}DeepSeek V4 Pro & 11.33\% & 30.83\% & 10.83\% & 6.67\% & 5.00\% & 3.33\% & 159.21M & 24.70h & \textcolor{black}{\$32.14} & 10.00\% & 30.00\% & 6.67\% & 5.83\% & 4.17\% & 3.33\% & 208.87M & 32.70h & \textcolor{black}{\$48.10} \\
\modellogo{glm}GLM 5.3 & 6.17\% & 26.67\% & 4.17\% & 0.00\% & 0.00\% & 0.00\% & 184.75M & 14.60h & \textcolor{black}{\$73.46} & 11.50\% & 27.50\% & 14.17\% & 9.17\% & 4.17\% & 2.50\% & 190.64M & 25.60h & \$80.50 \\

\addlinespace[2pt]
\bottomrule
\end{tabular}}
\end{table}

\begin{figure}[tbp]
  \centering
  \includegraphics[width=\linewidth]{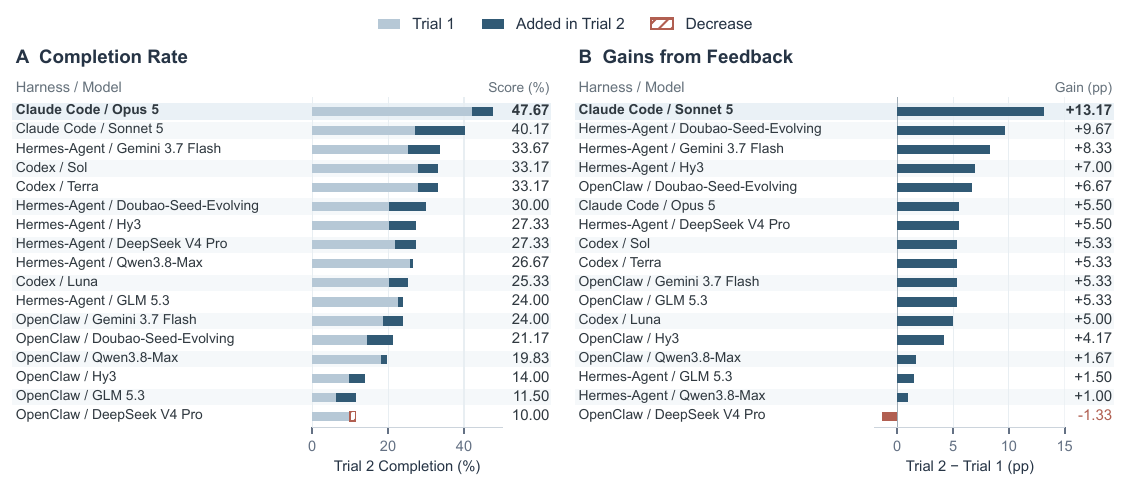}
  \par\vspace{-12pt}
  \caption{\textbf{Completion Rates and Feedback Gains.}}
  \label{fig:results-main-comparisons}
  \vspace{-10pt}
\end{figure}

\paragraph{Overall performance.}
Even the best-performing configuration, Claude Code with Opus~5, completes only 42.17\% of levels in Trial~1 and 47.67\% in Trial~2 (\Cref{fig:results-main-comparisons}A). In Trial~2, it completes all five levels in 28 of the 120 tasks. Across the 17 configurations, the median cumulative pass rate in Trial~2 is 62.50\% at L1 but only 8.33\% at L5 (\Cref{tab:main-results}). These results indicate that \benchname remains far from saturated for the evaluated models and agent harnesses.

\paragraph{Feedback and revision.}
Higher completion in Trial~1 does not necessarily imply larger absolute gains in Trial~2. Of the 17 configurations, 16 achieve higher overall completion in Trial~2, with a median increase of 5.33 percentage points (pp). Although Claude Code with Opus~5 leads in Trial~1 completion, Sonnet~5 shows a larger increase, at 13.17~pp compared with 5.50~pp (\Cref{fig:results-main-comparisons}A--B). These results motivate evaluating gains from the additional feedback and revision opportunity alongside performance under a single submission.

Improvements in overall completion do not always extend to later levels of the task ladder. With Sonnet~5, the number of fully completed tasks increases from 8 in Trial~1 to 23 in Trial~2. By contrast, both Qwen3.8-Max configurations show higher cumulative pass rates at L1--L3 but unchanged rates at L4--L5 (\Cref{tab:main-results}). These results motivate examining both how agents improve with feedback and revision and whether those gains extend to later assignments that build on earlier experience.

\begingroup
\titlespacing*{\paragraph}{0pt}{1.5ex plus 1pt minus 1pt}{1em}
\paragraph{Harness comparison.}
CLE evaluates the continual learning capability of the complete agent, including how its harness retains and reuses experience and how its model reflects on feedback. To examine differences across harnesses, we compare Hermes-Agent and OpenClaw on their six shared backbones. Hermes-Agent achieves higher completion for every backbone under both submission budgets (\Cref{fig:results-main-comparisons}A). Its mean advantage increases from 9.56~pp in Trial~1 to 11.42~pp in Trial~2, where their mean completion rates are 28.17\% and 16.75\%, respectively. The differences extend to gains from feedback and revision. With DeepSeek~V4~Pro, Hermes-Agent improves by 5.50~pp, whereas OpenClaw declines by 1.33~pp (\Cref{fig:results-main-comparisons}B). Appendix~\ref{app:additional-results} provides the detailed comparisons.

\begin{figure}[!htbp]
  \centering
  \includegraphics[width=\linewidth]{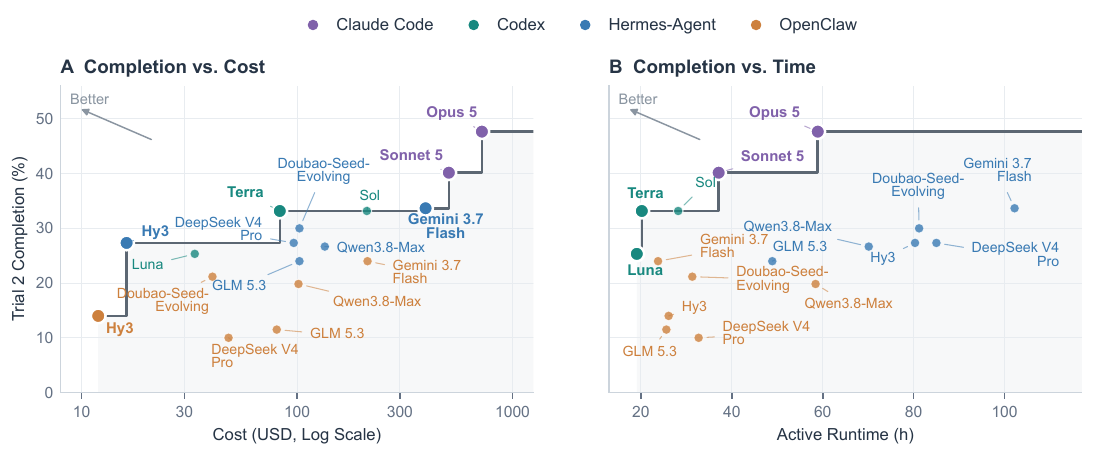}
  \par\vspace{-12pt}
  \caption{\textbf{Completion versus Cost and Runtime.} A--B show all 17 Trial~2 configurations; colours identify harnesses. Upper left is better. Bold labels and larger markers identify the observed Pareto frontier, traced by the step. Resource totals cover full evaluation runs.}
  \label{fig:results-main-resource-tradeoffs}
  \vspace{-10pt}
\end{figure}

\paragraph{Resource costs.}
Resource comparisons depend on both the completion metric and the resource considered. In Trial~2, Codex with Terra and Sol both achieve 33.17\% overall completion, but Terra costs \$83.16 and takes 20.20~hours, compared with \$210.83 and 28.20~hours for Sol. However, Sol completes 21 full tasks compared with Terra's 15 (\Cref{tab:main-results}). Equal overall completion thus masks differences in full-task completion, which matter when interpreting total resource use.

Monetary cost and runtime also yield different trade-offs. Hermes-Agent with Hy3 lies on the observed cost--completion frontier but not the runtime--completion frontier, achieving 27.33\% completion for \$16.18 while requiring 80.30~hours (\Cref{fig:results-main-resource-tradeoffs}A--B). These totals cover complete evaluation runs, whose workloads vary with the levels reached and revisions performed. Together, these results motivate assessing resource use alongside both overall and full-task completion, while treating monetary cost and runtime as separate constraints.
\par
\endgroup

{\color{black}
\hypersetup{allcolors=black}

}

\section{Related Work}
\label{sec:related}

\subsection[General Agent Capability Benchmarks]{General Agent Capability Benchmarks}
\label{subsec:related-environments}

General agent benchmarks evaluate capabilities such as reasoning, information gathering and synthesis, planning, and tool use. These capabilities are assessed through tasks ranging from question answering and software engineering to everyday workflows, with performance measured by answer correctness or successful task completion~\citep{mialon2023gaia,zhou2023webarena,xie2024osworld,jimenez2023swebench,zhao2025seecontrol}.

Professional workflows bring additional demands on task duration, environment continuity, and outcome verification. Agents' Last Exam covers expert-developed, long-horizon tasks with verifiable outcomes across multiple industries~\citep{sun2026agentslastexam}. ClawMark models multi-day workflows in which stateful services change between turns and checks the resulting service state with deterministic verifiers~\citep{meng2026clawmark}. WildClawBench runs tasks inside deployed command-line agent harnesses using real tools, combining rule-based checks, environment-state auditing, and model-based judgement to verify outcomes~\citep{ding2026wildclawbench}. These benchmarks characterise task-solving performance across diverse domains and workflow lengths. CLE evaluates task completion across connected professional assignments that allow experience to accumulate and inform later work.

\subsection[Memory and Context Learning Benchmarks]{Memory and Context Learning Benchmarks}
\label{subsec:related-memory}

Memory benchmarks assess the retention and use of information from interaction histories. LoCoMo evaluates long-term conversational memory through question answering, event summarisation, and multimodal dialogue generation~\citep{maharana2024locomo}. LongMemEval studies reasoning across sessions, tracking changing knowledge, and abstaining when historical evidence is insufficient~\citep{wu2025longmemeval}. LongMemEval-V2 evaluates memory of environment-specific experience through questions about web-agent state changes, workflows, and recurring failure modes, using evidence gathered from stored interaction trajectories~\citep{wu2026longmemevalv2}. CL-bench evaluates the acquisition and application of knowledge supplied in context through expert-authored tasks requiring new domain knowledge, rules, procedures, and patterns inferred from empirical data~\citep{dou2026clbench}. These evaluations address both access to prior information and the application of newly acquired knowledge. CLE evaluates knowledge reuse through professional task outcomes, while allowing each harness to determine what to retain and how to use it.

\subsection[Experience Accumulation and Reuse Benchmarks]{Experience Accumulation and Reuse Benchmarks}
\label{subsec:related-continual}

\begingroup
\setlength{\emergencystretch}{3em}
Recent work has introduced benchmarks that evaluate knowledge transfer, memory use, skill evolution, and adaptation over sequences of tasks. LifelongAgentBench constructs skill-grounded, interdependent tasks in database, operating-system, and knowledge-graph environments, with verifiable outcomes and sequential evaluation of knowledge transfer~\citep{zheng2025lifelongagentbench}. MemoryArena couples memory formation with later action through human-crafted, interdependent subtasks across web navigation, preference-constrained planning, information search, and formal reasoning~\citep{he2026memoryarena}. SkillFlow organises tasks into families with shared workflows: agents begin without stored skills, derive skill updates from execution trajectories and rubric feedback, and reuse the updated skills on subsequent tasks~\citep{zhang2026skillflow}. StuLife studies experience accumulation across stages of simulated university life, spanning enrolment, academic work, and personal development~\citep{caibenchmark}. SEA-Eval tracks success rates and token consumption over sequential task streams to characterise adaptation and efficiency~\citep{jiang2026sea}. CLE organises expert-authored professional assignments into five-level task ladders linked through tools and skills, data dependencies, strategies, user context, and domain knowledge. The evaluated harness determines how experience is retained and reused, while a common protocol controls opportunities for rubric-based diagnostic feedback and revision.\par
\endgroup

\paragraph{Comparison and Positioning of CLE.}

\Cref{tab:bench-cmp} compares representative settings along three complementary axes. \emph{Task design} describes the origin and form of the work; \emph{learning protocol} specifies progression, reuse, and revision opportunities; and \emph{feedback and grading} describe the signals available to the agent and the evidence supporting its score.
Definitions of the nine comparison criteria are provided in the table notes.
Across these dimensions, CLE combines professional task execution with experience reuse under an explicit learning protocol and auditable grading.

\begin{table}[p]
\centering
\caption{\textbf{Comparison of agent benchmarks.} \cmark{} indicates that the benchmark meets the criterion in that column. \protect\hmark{} indicates that it meets the criterion only partially or for some tasks or settings. \xmark{} indicates that the feature is absent or is not a focus of the benchmark.}
\label{tab:bench-cmp}

\renewcommand{\arraystretch}{1.10}
\setlength{\tabcolsep}{3pt}

\resizebox{\textwidth}{!}{%
\begin{tabular}{@{} >{\raggedright\arraybackslash}m{4.05cm} C C C B B B J M M @{}}
\toprule
\addlinespace[2pt]
& \multicolumn{3}{c}{\cellcolor{HeaderViolet}\textbf{Task Design}}
  & \multicolumn{3}{c}{\cellcolor{HeaderBlue}\textbf{Learning Protocol}}
  & \multicolumn{3}{c}{\cellcolor{HeaderRose}\textbf{Feedback and Grading}}\\
\cmidrule(l{2pt}r{2pt}){2-4}\cmidrule(l{2pt}r{2pt}){5-7}\cmidrule(l{2pt}r{2pt}){8-10}
\addlinespace[1pt]
\textbf{Benchmark}
& \makecell{\textbf{Human}\\\textbf{Authored /}\\\textbf{Verified}}
& \makecell{\textbf{Stateful}\\\textbf{Env.}}
& \makecell{\textbf{Multi-}\\\textbf{Modal}}
& \makecell{\textbf{Tiered}\\\textbf{Tasks}}
& \makecell{\textbf{Cross-}\\\textbf{level}\\\textbf{Reuse}}
& \makecell{\textbf{Config.}\\\textbf{Attempts}}
& \makecell{\textbf{Milestone}\\\textbf{Feedback}}
& \makecell{\textbf{Agentic}\\\textbf{Judge}}
& \makecell{\textbf{Audit}\\\textbf{Trail}}\\
\midrule
\addlinespace[2pt]

WebArena~\citep{zhou2023webarena}              & \cmark & \cmark & \xmark & \xmark & \xmark & \xmark & \xmark & \hmark & \hmark \\
OSWorld~\citep{xie2024osworld}                 & \cmark & \cmark & \cmark & \xmark & \xmark & \xmark & \xmark & \xmark & \hmark \\
Terminal-Bench~\citep{merrill2026terminalbench} & \cmark & \cmark & \cmark & \xmark & \xmark & \xmark & \xmark & \xmark & \cmark \\
GAIA~\citep{mialon2023gaia}                    & \cmark & \xmark & \cmark & \xmark & \xmark & \xmark & \xmark & \xmark & \hmark \\
CL-bench~\citep{dou2026clbench}                & \cmark & \xmark & \xmark & \hmark & \cmark & \xmark & \xmark & \hmark & \cmark \\
Agents' Last Exam~\citep{sun2026agentslastexam} & \cmark & \cmark & \cmark & \xmark & \xmark & \xmark & \xmark & \hmark & \cmark \\
Humanity's Last Exam~\citep{phan2025humanity}  & \cmark & \xmark & \cmark & \xmark & \xmark & \xmark & \xmark & \hmark & \cmark \\
AgentBench~\citep{liu2023agentbench}           & \hmark & \cmark & \xmark & \xmark & \xmark & \xmark & \hmark & \hmark & \hmark \\
SWE-bench~\citep{jimenez2023swebench}          & \hmark & \xmark & \xmark & \xmark & \xmark & \xmark & \xmark & \xmark & \cmark \\
$\tau$-bench~\citep{yao2024taubench}           & \cmark & \cmark & \xmark & \xmark & \xmark & \xmark & \xmark & \xmark & \cmark \\
MCPMark~\citep{wu2025mcpmark}                  & \cmark & \cmark & \xmark & \xmark & \xmark & \xmark & \xmark & \xmark & \hmark \\
Gaia2~\citep{froger2026gaia2}                  & \cmark & \cmark & \xmark & \hmark & \hmark & \xmark & \xmark & \hmark & \cmark \\
ARC-AGI-3~\citep{arcprize2026arcagi3}          & \cmark & \cmark & \cmark & \cmark & \cmark & \hmark & \cmark & \xmark & \cmark \\
Claw-Eval~\citep{ye2026claweval}               & \cmark & \cmark & \cmark & \xmark & \xmark & \xmark & \xmark & \hmark & \cmark \\
ClawBench~\citep{zhang2026clawbench}           & \cmark & \cmark & \cmark & \xmark & \xmark & \xmark & \xmark & \cmark & \cmark \\
ClawMark~\citep{meng2026clawmark}              & \cmark & \cmark & \cmark & \hmark & \cmark & \xmark & \xmark & \xmark & \cmark \\
WildClawBench~\citep{ding2026wildclawbench}    & \cmark & \cmark & \cmark & \xmark & \xmark & \xmark & \xmark & \hmark & \cmark \\
ClawArena~\citep{ji2026clawarena}              & \hmark & \cmark & \xmark & \hmark & \cmark & \xmark & \hmark & \xmark & \cmark \\
LifelongAgentBench \citep{zheng2025lifelongagentbench} & \hmark & \cmark & \xmark & \hmark & \hmark & \xmark & \xmark & \xmark & \hmark \\
MemoryArena~\citep{he2026memoryarena}
& \cmark & \hmark & \xmark
& \hmark & \cmark & \xmark
& \hmark & \hmark & \cmark \\
SkillFlow~\citep{zhang2026skillflow}           & \cmark & \cmark & \cmark & \hmark & \hmark & \xmark & \hmark & \xmark & \cmark \\
StuLife~\citep{caibenchmark}                   & \hmark & \cmark & \xmark & \hmark & \cmark & \xmark & \xmark & \xmark & \hmark \\

SEA-Eval~\citep{jiang2026sea}
& \cmark & \cmark & \cmark
& \hmark & \hmark & \xmark
& \xmark & \xmark & \xmark \\

\midrule
\addlinespace[1pt]
\rowcolor{OursYellow}
\textbf{\benchname(Ours)}
& \textbf{\cmark} & \textbf{\cmark} & \textbf{\cmark}
& \textbf{\cmark} & \textbf{\cmark} & \textbf{\cmark}
& \textbf{\cmark} & \textbf{\cmark} & \textbf{\cmark}\\
\addlinespace[2pt]
\bottomrule
\end{tabular}}
\par\smallskip
\begin{minipage}{\linewidth}
\footnotesize\raggedright
\setlength{\parskip}{3pt}
\noindent\textbf{Task design.}
\textbf{Human-authored / verified}: every task is human-authored or individually human-verified; \hmark{} marks mixed provenance such as generated tasks with only sampled human review.
\textbf{Stateful env.}: the agent interacts with a mutable web, app, API, OS, database, or service environment, rather than only local static files or final-answer strings; \hmark{} marks environments available for only part of the tasks.
\textbf{Multi-Modal}: at least one task includes a non-text input the agent must process, such as images, screenshots, audio, video, or visually grounded documents.

\noindent\textbf{Learning protocol.}
\textbf{Tiered tasks}: tasks unlock in an ordered curriculum; \hmark{} marks a fixed order without gating, or gating present only in a subset of tasks.
\textbf{Cross-level reuse}: later levels require artefacts, facts, or strategies produced in earlier levels; \hmark{} marks settings where earlier work only helps, or where the dependence appears only in a subset of tasks.
\textbf{Config.\ attempts}: the protocol grants a configurable per-level submission budget in which a failed attempt returns feedback the agent can act on before resubmitting.

\noindent\textbf{Feedback and grading.}
\textbf{Milestone feedback} reports progress through task stages without disclosing failures on individual evaluation criteria.
\textbf{Agentic judge} denotes grading by an evaluation agent with tools; \hmark{} indicates LLM-based grading without tool use.
\textbf{Audit trail} provides access to judgements on individual criteria, supporting evidence, or evaluator logs; \hmark{} indicates task-level pass/fail results with execution logs only.
\end{minipage}
\end{table}

\section{Conclusion}
\label{sec:conclusion}
We introduced the Continual Learning Exam (\benchname), a benchmark for evaluating continual learning in realistic human--AI cowork settings. \benchname comprises 120 human-authored and human-verified tasks, each organised as a five-level ladder that connects successive assignments through reusable information and artefacts. Stateful environments and human-defined rubrics support task execution and verification throughout each ladder. Across 17 agent configurations with fixed model parameters, the best-performing system completes only 47.67\% of levels, even with diagnostic feedback and an opportunity for revision. Feedback and revision improve most configurations, but these gains do not consistently extend to completing entire task ladders. \benchname remains challenging for current models and agent harnesses, providing a setting for developing and evaluating methods that acquire, retain, and reuse experience to improve performance across successive assignments.

\FloatBarrier
{\small\bibliographystyle{plainnat}\bibliography{cle_refs}}

\newpage
\appendix
\raggedbottom
\Needspace{5\baselineskip}
\section{Authors and Affiliations}
\label{app:authors-affiliations}

\noindent Haoyu Zhao$^{*}$, University College London\\[2pt]
Sitian Tu$^{*}$, University College London\\[2pt]
Yuxuan Huang$^{*}$, University of Liverpool\\[2pt]
Huichi Zhou, University College London\\[2pt]
Zhangyue Yin, University College London\\[2pt]
Jiahao Huang, Imperial College London\\[2pt]
Jingyuan Xu, Imperial College London\\[2pt]
Tianmiao Shi, University College London\\[2pt]
Jin Shi, University College London\\[2pt]
Nianxi Zhang, University College London\\[2pt]
Yangmiao Zhang, University of Electronic Science and Technology of China\\[2pt]
Zhuolang Chan, University College London\\[2pt]
Huiyan Hu, University College London\\[2pt]
Chenchao Sheng, University College London\\[2pt]
Yaqing Luo, Imperial College London\\[2pt]
Yue Chen, Peking University\\[2pt]
Huaxiu Yao, University of North Carolina at Chapel Hill\\[2pt]
Meng Fang, University of Liverpool\\[2pt]
Jun Wang$^{\dagger}$, University College London\par

\smallskip
\noindent $^{*}$ Equal contribution. $^{\dagger}$ Corresponding author: \href{mailto:jun.wang@cs.ucl.ac.uk}{jun.wang@cs.ucl.ac.uk}.

\section[Comparison of Related Learning Concepts]{Comparison of Related Learning Concepts}
\label{app:related-concepts}
\begingroup
\setlength{\emergencystretch}{2em}

This appendix provides the full comparison of learning paradigms and mechanisms summarised in \Cref{subsec:cl-related-concepts}. The notation follows \Cref{eq:state-update} and the parameter and state updates defined in that subsection.

{
  \small
  \renewcommand{\arraystretch}{1.13}
  \setlength{\tabcolsep}{4pt}
  \setlength{\LTcapwidth}{\linewidth}
  \begin{longtable}{@{}>{\raggedright\arraybackslash}p{0.20\linewidth}>{\raggedright\arraybackslash}p{0.44\linewidth}>{\raggedright\arraybackslash}p{0.32\linewidth}@{}}
    \caption{\textbf{Related Perspectives on Continual Learning.} Selected learning paradigms and mechanisms, their relation to CL, and representative updates in \Cref{eq:state-update}. Arrows indicate update targets; equality indicates a fixed component. The memory, reflection, and skill entries show frozen-backbone examples.}
    \label{tab:notions-full} \\
    \toprule
    \rowcolor{HeaderViolet!65}
    \textbf{Concept} & \leavevmode{}\textbf{Perspective on Continual Learning} & \leavevmode{}\textbf{Update in $(\theta_n,s_n)$} \\
    \midrule
    \endfirsthead
    \multicolumn{3}{@{}l}{\tablename~\thetable{} (continued)} \\
    \toprule
    \rowcolor{HeaderViolet!65}
    \textbf{Concept} & \leavevmode{}\textbf{Perspective on Continual Learning} & \leavevmode{}\textbf{Update in $(\theta_n,s_n)$} \\
    \midrule
    \endhead
    \midrule
    \multicolumn{3}{r@{}}{\textit{Continued on next page}} \\
    \endfoot
    \bottomrule
    \endlastfoot
    \multicolumn{3}{@{}l}{\textbf{Learning Settings and Paradigms}} \\*
    \addlinespace[2pt]
    \leavevmode{}Continual Learning & \leavevmode{}Acquiring, refining, retaining, and applying knowledge from experience to improve later capabilities while preserving useful prior knowledge~\citep{ring1994continual,parisi2019continual}. & \leavevmode{}$\theta_n\rightarrow\theta_{n+1}$ and/or $s_n\rightarrow s_{n+1}$ through $U_n$. \\
    \addlinespace[2pt]
    \leavevmode{}Lifelong Learning & \leavevmode{}Emphasises the lifetime over which CL accumulates and transfers knowledge; often used interchangeably with CL~\citep{thrun1995lifelong,parisi2019continual}. & \leavevmode{}$\theta_n\rightarrow\theta_{n+1}$ and/or $s_n\rightarrow s_{n+1}$ across a continuing stream of experience. \\
    \addlinespace[2pt]
    Online Learning & Learns through sequential prediction, feedback, and update, typically aiming to minimise cumulative loss~\citep{shalevshwartz2012online}. In CL, online updates combine adaptation to incoming experience with retention of useful prior knowledge. & $\theta_n\rightarrow\theta_{n+1}$ and/or $s_n\rightarrow s_{n+1}$ as experience $e_n$ arrives. \\
    \addlinespace[2pt]
    Transfer Learning & Uses knowledge from source tasks or domains to improve learning or performance on target tasks or domains~\citep{pan2010transfer}. In CL, knowledge transfer can benefit both subsequent and previously learned tasks~\citep{lopezpaz2017gem}. & Knowledge in $\theta_n$ or $s_n$ is reused and may be refined through $U_n$ to support other tasks. \\
    \addlinespace[2pt]
    \leavevmode{}Multi-Task Learning & \leavevmode{}Jointly learns multiple tasks through shared representations or parameters. In CL, joint training on current and replayed experience can support knowledge sharing and retention~\citep{caruana1997multitask,shin2017generativereplay}. & \leavevmode{}$\theta_n\rightarrow\theta_{n+1}$ using multi-task experience $e_n$; $s_n$ may retain examples or task information. \\
    \addlinespace[2pt]
    Curriculum Learning & Organises learning examples or tasks in a purposeful sequence, often from easier to harder~\citep{bengio2009curriculum}. In CL, a curriculum can help later learning build on earlier knowledge. & Determines the sequence or weighting of experience $e_n$ supplied to $U_n$; updates can affect $\theta_n$, $s_n$, or both. \\
    \addlinespace[2pt]
    Meta-Learning & \leavevmode{}Optimises how a learner adapts using experience across tasks. In CL, learned initialisations, representations, or update rules can improve adaptation while reducing interference with earlier knowledge~\citep{finn2017maml,andrychowicz2016learning,javed2019metalearning}. & \leavevmode{}$\theta_n\rightarrow\theta_{n+1}$ and/or $s_n\rightarrow s_{n+1}$; retained initialisations or control state improve later adaptation. \\
    \addlinespace[4pt]
    \multicolumn{3}{@{}l}{\textbf{Update and Experience-Reuse Mechanisms}} \\*
    \addlinespace[2pt]
    \leavevmode{}Parametric Learning & \leavevmode{}Incorporates knowledge from data and learning signals into model weights. CL aims to accumulate knowledge across updates while preserving useful earlier capabilities~\citep{xi2024agentgym,xi2025agentgymrl}. & \leavevmode{}$\theta_n\rightarrow\theta_{n+1}$ through $U_n^\theta$; $s_n$ may retain replay data or learning statistics. \\
    \addlinespace[2pt]
    In-Context Learning & \leavevmode{}Applies demonstrations or prior interactions through context. It supports CL when relevant experience is retained and brought into later contexts~\citep{brown2020fewshot}. & \leavevmode{}$\theta_{n+1}=\theta_n$; $s_n\rightarrow s_{n+1}$ when context or supporting memory carries experience forward. \\
    \addlinespace[2pt]
    Test-Time Training & \leavevmode{}Optimises a learning objective on test-time inputs to update model weights or adaptive memory. Applications include test-time adaptation (TTA) to distribution shifts~\citep{wang2021tent} and long-context sequence modelling. Retained adaptations carry knowledge into subsequent updates and predictions~\citep{sun2020testtime,sun2025testtimelayers,tandon2025ttte2e}. & \leavevmode{}$\theta_n\rightarrow\theta_{n+1}$ when adapting model weights, or $s_n\rightarrow s_{n+1}$ for retained adaptive memory. \\
    \addlinespace[2pt]
    \leavevmode{}Test-Time Scaling & \leavevmode{}Allocates additional inference computation for search, checking, or reflection~\citep{snell2024scaling}. Retained solutions or lessons can supply experience for later learning updates. & \leavevmode{}Without retention: $\theta_{n+1}=\theta_n$ and $s_{n+1}=s_n$.\par With retained outcomes: $s_n\rightarrow s_{n+1}$. \\
    \addlinespace[2pt]
     \leavevmode{}Memory & \leavevmode{}Carries experience beyond the current interaction and retrieves relevant knowledge for later work, supplying persistent state for CL~\citep{packer2023memgpt,zhou2025memento}. Retrieval-augmented generation (RAG) conditions generation on retrieved records~\citep{lewis2020rag}; updates to that store can make new experience available to later interactions. & \leavevmode{}$\theta_{n+1}=\theta_n$; $s_n\rightarrow s_{n+1}$ through memory writing, revision, and consolidation. \\
    \addlinespace[2pt]
    \leavevmode{}Case-Based Reasoning & \leavevmode{}Accumulates concrete cases and adapts relevant past solutions to new situations; evaluating and retaining new cases extends knowledge for subsequent work~\citep{aamodt1994cbr,zhou2025memento}. & \leavevmode{}$s_n\rightarrow s_{n+1}$ updates the case store and any learned retrieval state; the backbone can remain fixed. \\
    \addlinespace[2pt]
    \leavevmode{}Reflection & \leavevmode{}Uses feedback to evaluate earlier behaviour, extract lessons, and revise knowledge for later decisions. Retaining and reusing these lessons supports CL~\citep{shinn2023reflexion,zhou2025memento,wang2025memento2}. & \leavevmode{}$\theta_{n+1}=\theta_n$; $s_n\rightarrow s_{n+1}$ through $U_n^s$, retaining revised lessons or strategies. \\
    \addlinespace[2pt]
    Skill Learning & \leavevmode{}Acquires and refines reusable procedures from experience. In CL, these procedures can be retained, composed, and adapted for later tasks~\citep{wang2023voyager,wang2024awm,zhou2026memento}. & \leavevmode{}$\theta_{n+1}=\theta_n$; $s_n\rightarrow s_{n+1}$ adds or revises executable procedures. \\
    \addlinespace[4pt]
    \multicolumn{3}{@{}l}{\textbf{Self-Improvement Frameworks}} \\*
    \addlinespace[2pt]
    \leavevmode{}Harness Self-Evolution & \leavevmode{}Uses accumulated feedback to revise how the agent acts and manages experience, carrying improvements in the harness into later assignments~\citep{zhang2025darwin,agrawal2025gepa}. & \leavevmode{}$s_n\rightarrow s_{n+1}$ changes prompts, tools, memory-management rules, or control code; $\theta_n$ can remain fixed. \\
    \addlinespace[2pt]
    Recursive Self-Improvement (RSI) & \leavevmode{}Retains self-modifications that improve how the system generates, evaluates, or implements further improvements~\citep{yudkowsky2007levels,zhao2026memento3}. In experience-driven RSI, accumulated knowledge can improve the learning mechanism itself. & \leavevmode{}$\theta_n\rightarrow\theta_{n+1}$ and/or $s_n\rightarrow s_{n+1}$; revised learning procedures in $s_{n+1}$ can change the effective $U_{n+1}$. \\
    \addlinespace[2pt]
    \leavevmode{}G\"odel Machine & \leavevmode{}A formal route to self-improvement in which code changes require a proof of greater expected utility. Verified revisions can preserve and extend the system's learning machinery~\citep{schmidhuber2003godel}. & \leavevmode{}$s_n\rightarrow s_{n+1}$ can rewrite learning and proof-search code; represented model parameters may also change. \\
  \end{longtable}
}
\endgroup

\section{Agent Auto Review with Feedback}
\label{app:auto-review}

We use a task-review agent to conduct the initial review of each task draft and provide feedback to its author for revision before final review by two human reviewers.

\paragraph{Human expertise as a review skill.}
The auto-review process follows criteria derived from human expertise in task design. We consolidate review criteria developed and iteratively refined by the benchmark development team into a reusable review skill. The skill specifies what the review agent should inspect, how to apply the review criteria, and how to organise its findings into actionable feedback for the task contributor.

\begin{figure}[htbp]
\centering
  \includegraphics[width=\textwidth]{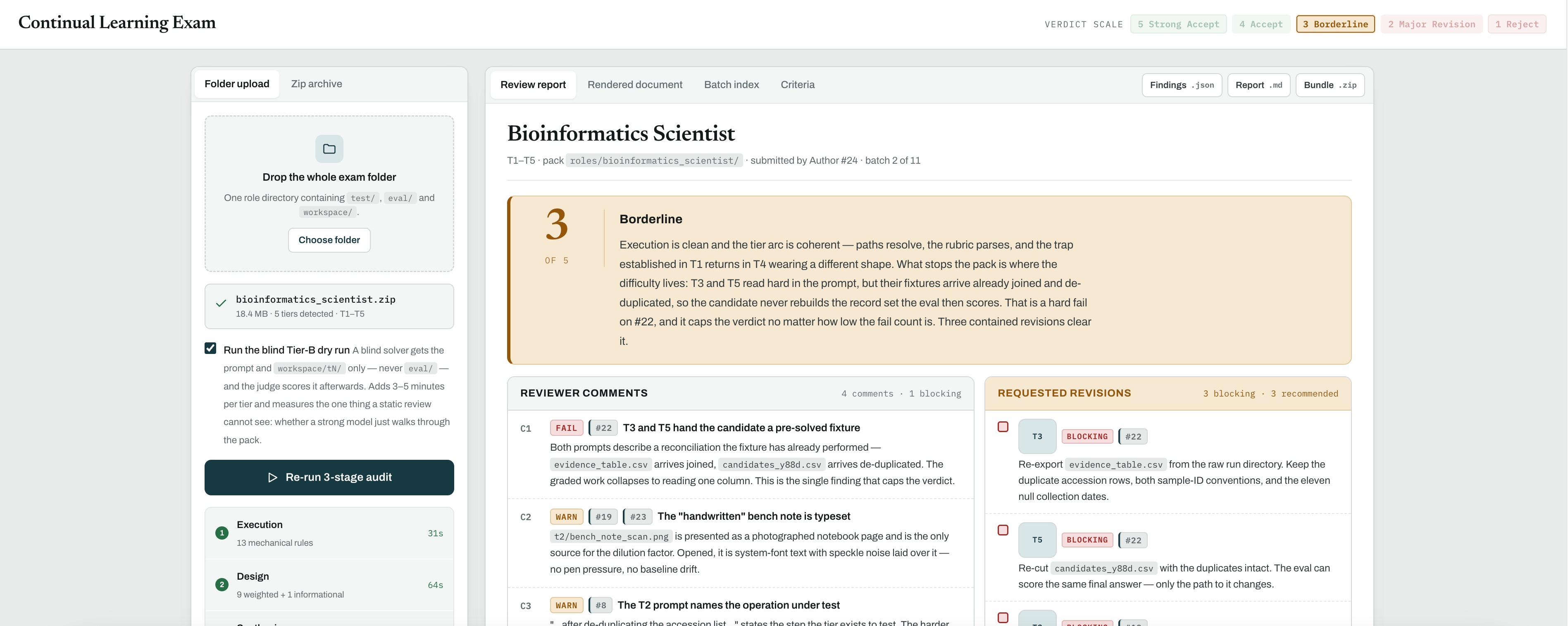}
  \caption{\textbf{Review-Agent Interface Used during Task Construction.} A task contributor uploads a complete five-level task together with its supporting files and receives a conference-style verdict, review comments, and revision suggestions after the review agent completes its execution and design checks.}
  \label{fig:review-agent-interface}
\par
\end{figure}

\paragraph{Feedback-driven human--agent iteration.}
As illustrated in \Cref{fig:construction,fig:review-agent-interface}, an author submits a draft and the review agent applies the skill to the full ladder, including its instructions, artefacts, environment state, and evaluation rubric. It returns evidence-grounded findings that identify the affected component, explain why it violates a design criterion, and suggest a concrete direction for revision. Task authors use the feedback to revise their drafts and submit them for another round of review. This cycle can repeat throughout task construction.

\paragraph{Final human review.}
Tasks receiving an automated verdict of Accept or Strong Accept proceed to independent review by two human reviewers. A task is admitted to the benchmark only if both reviewers accept it; otherwise, it returns to the author for revision. This separation of roles combines the coverage and consistency of agent-assisted review with the contextual judgement and accountability of human experts. The development team can refine the review skill based on recurring issues identified during task construction, allowing lessons from earlier review rounds to improve feedback for later task authors.

\section{Human--Agent Grading Agreement}
\label{app:human-agent-agreement}

\paragraph{Alignment with human grading.}
We assess the evaluation agent's agreement with human grading at the rubric-item
level. The validation sample covers all five levels of 12 tasks, comprising
60 submissions and 732 rubric items. Human annotators
independently label each item and resolve disagreements to establish reference
labels. All judges are provided with the same fixed submissions and environment
evidence. Our evaluation agent with DeepSeek~V4~Flash achieves 96.86\% agreement
with the human reference labels. \Cref{tab:human-agent-agreement} reports
pairwise agreement among the human reference labels and the three agent judges.

\begin{table}[htbp]
\centering
  \caption{\textbf{Agreement on Individual Rubric Items.}
  Each cell reports the percentage of rubric items receiving the same label
  from the corresponding pair of judges. Column numbers correspond to the
  numbered rows. Human annotators independently assess each rubric item and
  resolve disagreements to establish the reference labels. Judge 2 is
  \benchname's evaluation agent with DeepSeek~V4~Flash. Judges 3 and 4 provide
  additional comparisons using representative frontier agent--model
  combinations: Claude Code with Claude Opus 5 and Codex with GPT-5.6-Sol,
  respectively.}
  \label{tab:human-agent-agreement}

  \renewcommand{\arraystretch}{1.25}
  \setlength{\tabcolsep}{15pt}
  \small
  \begin{tabular}{@{}lcccc@{}}
    \toprule
    & \multicolumn{4}{c}{Judge} \\
    \cmidrule(l){2-5}
    & 1 & 2 & 3 & 4 \\
    \midrule
    1 \enspace Human Annotators             & 100.00\%  & 96.86\% & 97.27\% & 95.49\% \\
    2 \enspace Ours w/ DeepSeek V4 Flash    & 96.86\% & 100.00\%  & 96.58\% & 92.90\% \\
    3 \enspace Claude Code w/ Claude Opus 5 & 97.27\% & 96.58\% & 100.00\%  & 95.22\% \\
    4 \enspace Codex w/ GPT-5.6-Sol         & 95.49\% & 92.90\% & 95.22\% & 100.00\%  \\
    \bottomrule
  \end{tabular}
\par
\end{table}

\paragraph{Stability across repeated runs.}
To measure repeatability, we grade a fixed set of submissions covering 556 rubric
items three times using the evaluation agent with DeepSeek~V4~Flash.
Each run starts from a fresh session, with the model, tool access,
rubrics, and submitted evidence held fixed. As shown in
\Cref{tab:deepseek-judge-stability}, 553 of the 556 items receive the same label
in all three runs, yielding 99.46\% agreement. These results show high repeatability of item-level grading
under fixed evaluation conditions.

\begin{table}[htbp]
\centering
  \caption{\textbf{Stability of the evaluation agent across repeated runs.}
  A fixed set of submissions is graded in three fresh sessions.
  Agreement is the percentage of the 556 rubric items that receive the same label in all three runs.}
  \label{tab:deepseek-judge-stability}

  \renewcommand{\arraystretch}{1.25}
  \setlength{\tabcolsep}{10pt}
  \small
  \begin{tabular}{@{}lccc@{}}
    \toprule
    Judge & \makecell{Total checklist\\items} & \makecell{Consistent\\items} & \makecell{Agreement\\(\%)} \\
    \midrule
    Ours w/ DeepSeek V4 Flash & 556 & 553 & 99.46\\
    \bottomrule
  \end{tabular}
\par
\end{table}

\section{Additional Benchmark Data Analysis}
\label{app:data-analysis}

\paragraph{Level-wise statistics.} The average checklist size increases from 18.74 items at $L_1$ to 36.83 at $L_5$, task-brief length from 246 to 466 words, and the number of rubric sections from 4.10 to 5.74 (\Cref{tab:level-profile}). Increases in checklist size and task-brief length are larger at the upper levels. Since a submission must satisfy every checklist item to pass, later levels require more conditions to be met simultaneously.

\begin{table}[htbp]
\centering
\hypersetup{allcolors=black}\color{black}
\caption{\textbf{Task Statistics by Level.} Each column summarises checklist size, task-brief length, and the number of rubric sections across the 120 tasks at the corresponding level.}
\label{tab:level-profile}

\renewcommand{\arraystretch}{1.15}
\begin{tabular}{@{}lccccc@{}}
\toprule
& \textbf{\textcolor{black}{$L_1$}} & \textbf{\textcolor{black}{$L_2$}} & \textbf{\textcolor{black}{$L_3$}} & \textbf{\textcolor{black}{$L_4$}} & \textbf{\textcolor{black}{$L_5$}}\\
\midrule
Checklist Items (Mean)   & 18.74 & 22.07 & 25.83 & 30.90 & 36.83\\
Checklist Items (Median) & 16.00 & 19.00 & 22.00 & 27.50 & 34.50\\
Checklist Items (Max)    & 75   & 85   & 82   & 87   & 112\\
Brief Length (Mean Words)     & 246  & 271  & 300  & 375  & 466\\
Rubric Sections (Mean)   & 4.10 & 4.39 & 4.52 & 5.21 & 5.74\\
\bottomrule
\end{tabular}
\par
\end{table}

\begin{figure}[tb]
\centering
  \includegraphics[width=\linewidth,trim=0 30bp 0 0,clip]{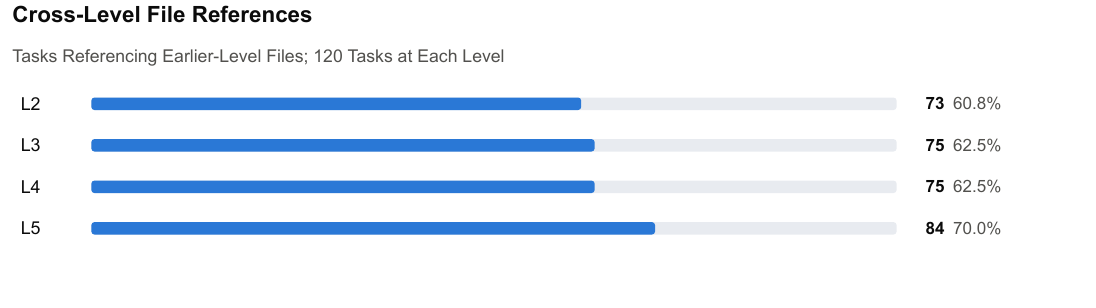}
  \caption{\textbf{Cross-Level File References.} Number and percentage of the 120 task ladders whose task specification or evaluation rubric at each level refers to files from an earlier level.}
  \label{fig:structure}
\par
\end{figure}

\paragraph{Cross-level file references.} References to earlier-level files are common across CLE and become more prevalent at later levels. Overall, 88 of the 120 task ladders contain at least one such reference in a task specification or evaluation rubric. The number of ladders with these references increases from 73 at $L_2$ to 84 at $L_5$ (\Cref{fig:structure}).

\section{Additional Analysis}
\label{subsec:analysis}
\label{subsec:preceding-levels}

\label{app:additional-results}

\subsection{Task Progression}
\begin{figure}[htbp]
\centering
  \includegraphics[width=\linewidth]{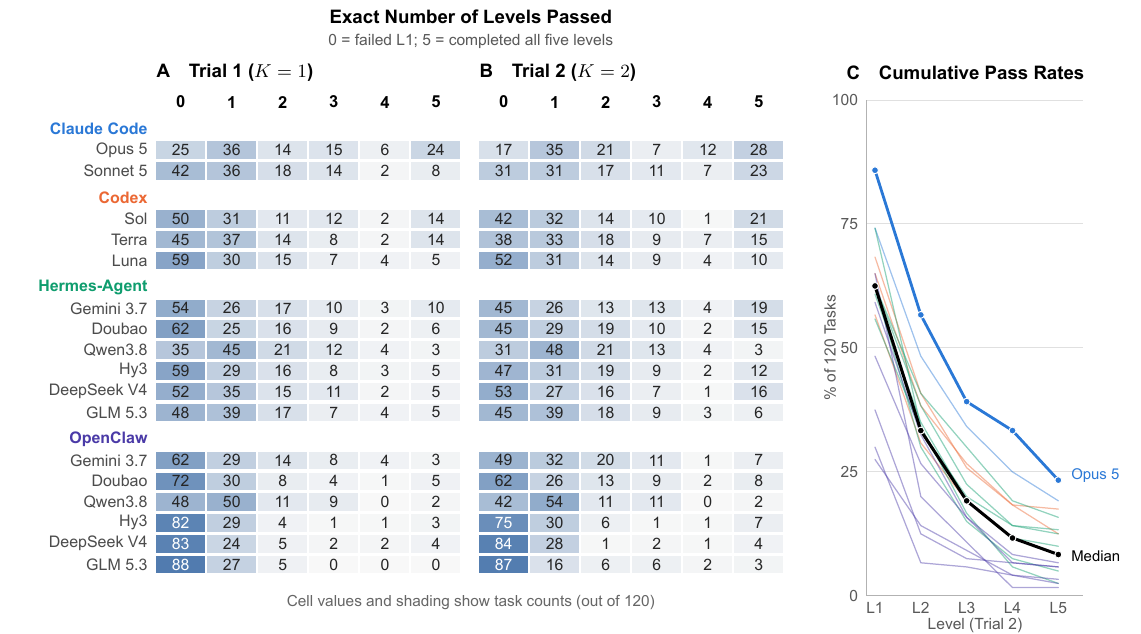}
  \caption{\textbf{Task Progression across Five Levels.} \textbf{(A--B)} Distribution of tasks by the number of levels completed (0--5) in Trial~1 ($K=1$) and Trial~2 ($K=2$). Each row sums to 120 tasks; darker cells indicate larger counts. \textbf{(C)} Cumulative pass rates in Trial~2, measured over all 120 tasks at each level. Colours identify harnesses. The black curve shows the median across the 17 agent configurations. Claude Code with Opus~5, the configuration with the highest overall completion, is highlighted.}
  \label{fig:results-depth}
\end{figure}

Early termination remains common under both submission budgets (\Cref{fig:results-depth}A--B). For every configuration in Trial~1, more than half of the 120 tasks end with zero or one completed level, from 61 tasks (50.83\%) for Claude Code with Opus~5 to 115 (95.83\%) for OpenClaw with GLM~5.3. Across the 17 configurations, the median number of such tasks falls only from 87 (72.50\%) in Trial~1 to 80 (66.67\%) in Trial~2, and Claude Code with Opus~5 is the only configuration that drops below half in Trial~2, at 52 tasks (43.33\%).

Performance at L1 constrains but does not determine how far a configuration gets (\Cref{fig:results-depth}C). In Trial~2, the Spearman correlation across the 17 configurations between the L1 pass rate and overall completion is $\rho=0.75$, but between the L1 and L5 pass rates it is only $\rho=0.42$. The two Qwen3.8-Max configurations are the clearest case. At L1 they pass 74.17\% of tasks with Hermes-Agent and 65.00\% with OpenClaw, both above the median of 62.50\%. At L5 they pass 2.50\% and 1.67\%, both below the median of 8.33\%, and the OpenClaw rate is the lowest of the 17 configurations. OpenClaw falls below the median at L2 and Hermes-Agent at L3.

\begin{figure}[htbp]
\centering
  \includegraphics[width=\linewidth]{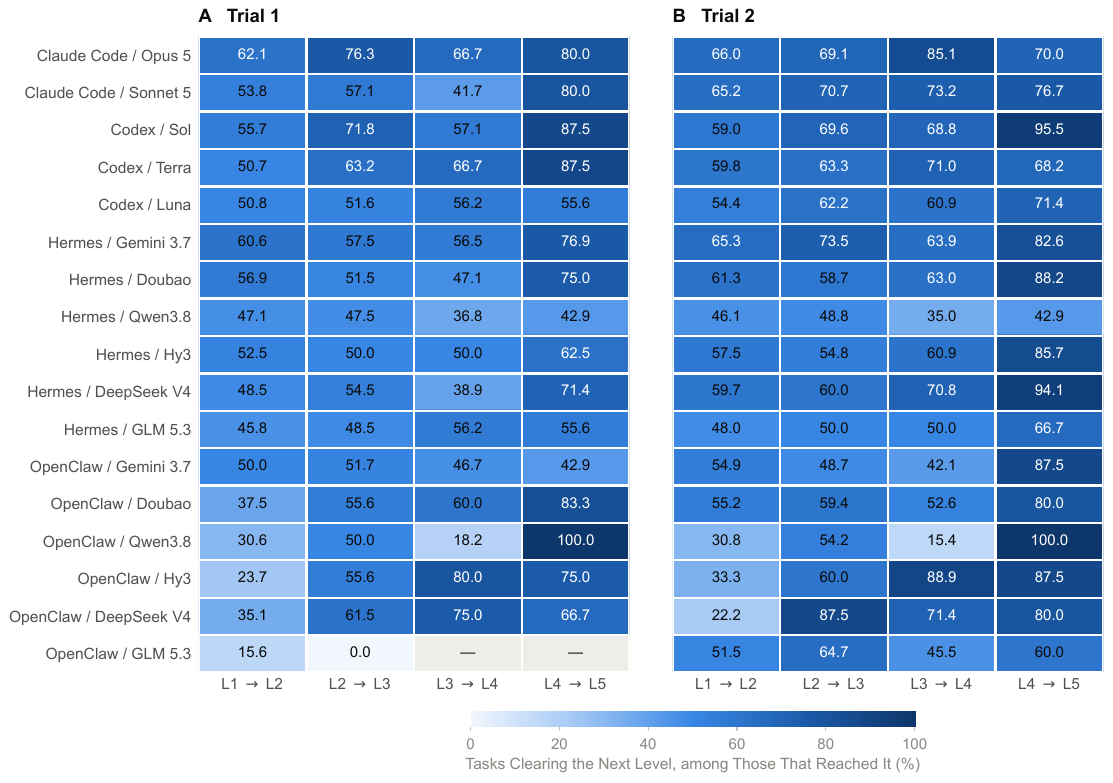}
  \caption{\textbf{Conditional Progression between Adjacent Levels.} Each cell shows the percentage of tasks that passed $L_{k+1}$ among those that passed $L_k$. Values are rounded to one decimal place. A dash indicates that no task passed $L_k$.}
  \label{fig:results-conditional}
\end{figure}

\begingroup
\looseness=-1
The median conditional pass rate across configurations rises at every transition (\Cref{fig:results-conditional}), from 50.00\% (L1$\to$L2) to 75.00\% (L4$\to$L5) in Trial~1 and from 55.17\% to 80.00\% in Trial~2. In Trial~1, the last two medians exclude OpenClaw with GLM~5.3, for which no task passed L3. These increases do not show that later assignments are easier. Each rate is measured only on tasks that cleared every earlier level, so it is computed over a progressively more favourable subset of tasks, and on those tasks the agent also carries forward the experience of the levels it has already cleared. The conditional pass rates reflect both effects.
\par
\endgroup

Despite the rising medians, later bottlenecks remain common. In Trial~2, 8 of the 17 configurations have a later transition with a lower conditional pass rate than their L1$\to$L2 rate. OpenClaw with Qwen3.8-Max, for example, passes L4 on only 2 of the 13 tasks that passed L3 (15.38\%). These late rates also rest on few tasks. Across configurations, the median denominator in Trial~2 falls from 75 tasks at L1$\to$L2 to 14 at L4$\to$L5, and 7 of the 17 configurations have fewer than 10 tasks at the final transition. The same OpenClaw configuration's L4$\to$L5 rate is 100.00\%, from 2 tasks.

\subsection{Feedback and Revision Gains}
\begin{figure}[htbp]
\centering
  \includegraphics[width=\linewidth]{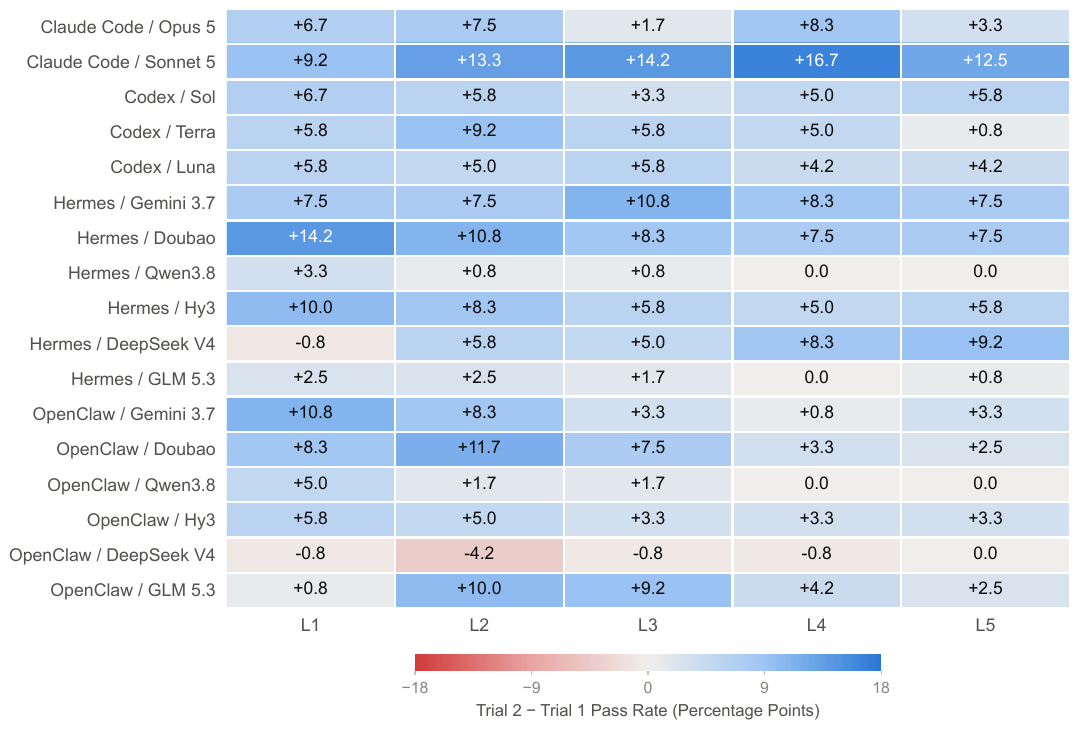}
  \caption{\textbf{Feedback and Revision Gains at Each Level.} Cells show Trial~2 minus Trial~1 pass rate in percentage points across all 120 tasks. Blue indicates an increase, red a decrease, and the neutral midpoint no change. Differences use integer pass counts before rounding.}
  \label{fig:results-gains}
\end{figure}

Small absolute gains at later levels can represent substantial improvements over their low baselines. Across configurations, the median increase from Trial~1 to Trial~2 is 7.50~pp at L2 and 3.33~pp at L5 (\Cref{fig:results-gains}). Dividing each configuration's increase by its own Trial~1 pass rate before taking the median, the corresponding median relative increases are 22.50\% and 75.00\%; the L2 median uses all 17 configurations and the L5 median 16, excluding OpenClaw with GLM~5.3, whose Trial~1 L5 pass rate is zero.

Changes in first-level performance alone do not capture the gains in full-task completion. For Claude Code with Sonnet~5, gains reach $+16.67$~pp at L4 and $+12.50$~pp at L5, compared with $+9.17$~pp at L1. Hermes-Agent with DeepSeek~V4~Pro loses 0.83~pp at L1 while gaining 9.17~pp at L5, showing that more tasks can be completed in full even when fewer pass L1. Each difference compares two complete gated runs, so it reflects both access to later levels and success on the levels reached.

\subsection{Harness Comparison}
\begin{figure}[htbp]
\centering
  \includegraphics[width=\linewidth]{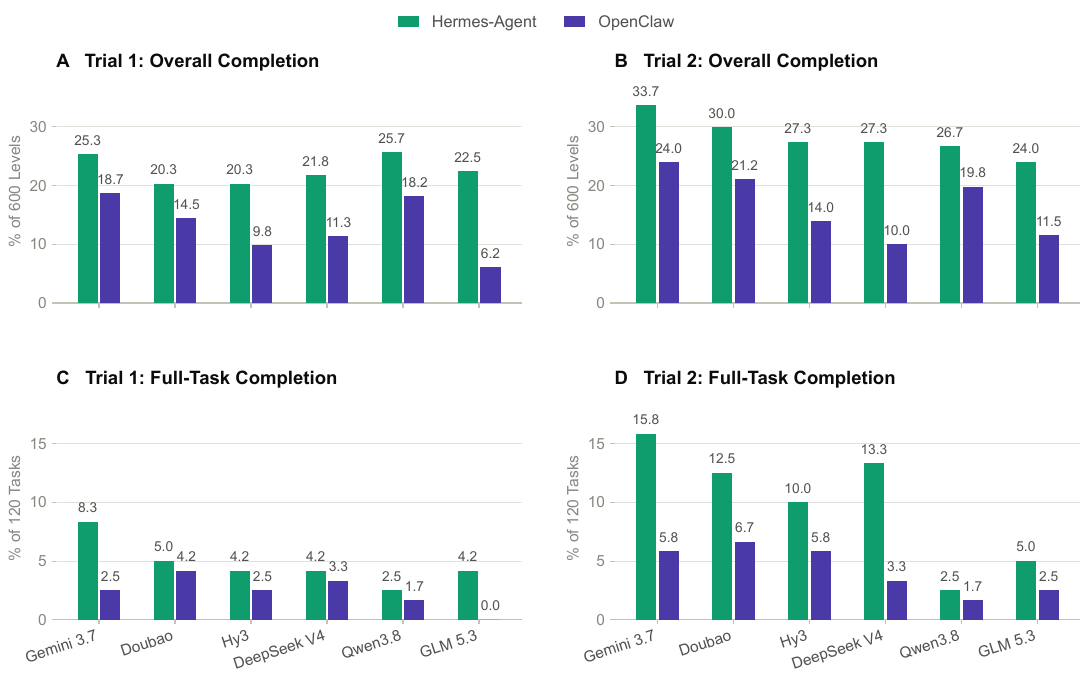}
  \caption{\textbf{Harness Comparison on Six Shared Backbones.} Grouped bars compare Hermes-Agent and OpenClaw under the same submission budget. \textbf{(A--B)} Overall completion out of 600 levels. \textbf{(C--D)} Full-task completion out of 120 tasks.}
  \label{fig:results-harness}
\end{figure}

Hermes-Agent achieves higher full-task completion than OpenClaw for every shared backbone under both submission budgets (\Cref{fig:results-harness}). The margin is only one task in four of the twelve backbone--budget comparisons. Its mean advantage in full-task completion across the six backbones grows from 2.36~pp in Trial~1 to 5.56~pp in Trial~2. Its Trial~2 advantage in overall completion ranges from 6.83~pp for Qwen3.8-Max to 17.33~pp for DeepSeek~V4~Pro. In Trial~2, its mean actor-token usage and cost are 1.90 and 1.71 times those of OpenClaw, respectively (\Cref{tab:main-results}). These totals cover runs that reach different levels and perform different amounts of revision, so they do not establish efficiency on matched workloads.

\subsection{Effect of Model Generation}

To examine whether newer model generations improve performance on \benchname, we compare Gemini~3.6, 3.7, and 3.8 Flash with Hermes-Agent under both submission budgets (\Cref{tab:model-generations}). Across generations, we hold the harness configuration, task set and order, tools, and state retention fixed.

Gemini~3.7 Flash achieves the highest overall completion and full-task completion in both trials. In Trial~1, its overall completion reaches 25.33\%, compared with 21.00\% for Gemini~3.6 and 21.17\% for Gemini~3.8. Gemini~3.8 also consumes more actor tokens than Gemini~3.7 (795.78M versus 486.25M), while completing fewer levels and full tasks.

All three models improve in Trial~2. Gemini~3.8 shows the largest increase in overall completion, gaining 9.50 percentage points, but still trails Gemini~3.7 in both overall completion (30.67\% versus 33.67\%) and fully completed tasks (14 versus 19). Across these three generations, performance therefore does not improve consistently with model updates, even when diagnostic feedback and revision are available.

\begin{table}[htbp]
\centering
\hypersetup{allcolors=black}
\color{black}
\caption{\color{black}\textbf{Gemini model-generation comparison with Hermes-Agent.} All variants use the full 120-task protocol and the metrics of \Cref{tab:main-results}. Gemini~3.7 Flash reuses the main-table runs.}
\label{tab:model-generations}
\newlength{\GenCompletionWidth}\settowidth{\GenCompletionWidth}{zhangyue}
\newlength{\GenPassWidth}\settowidth{\GenPassWidth}{00.00\%}
\newlength{\GenTokensWidth}\settowidth{\GenTokensWidth}{486.25M}
\newlength{\GenTimeWidth}\settowidth{\GenTimeWidth}{95.50h}
\newlength{\GenCostWidth}\settowidth{\GenCostWidth}{\$215.64}
\renewcommand{\arraystretch}{1.3}
\setlength{\tabcolsep}{3.5pt}
\resizebox{\textwidth}{!}{%
\begin{tabular}{@{} l *{18}{c} @{}}
\toprule
\addlinespace[2pt]
\multirow{3}{*}{\textbf{Model}} & \multicolumn{9}{c}{\textbf{Trial 1}} & \multicolumn{9}{c}{\textbf{Trial 2}} \\
\cmidrule(l{2pt}r{2pt}){2-10}\cmidrule(l{2pt}r{2pt}){11-19}
 & \multirow{2}{*}{\makebox[\GenCompletionWidth]{\textbf{Compl.}}} & \multicolumn{5}{c}{\textbf{Per-Level Pass}} & \multirow{2}{*}{\makebox[\GenTokensWidth]{\textbf{\makecell{Actor\\Tok.}}}} & \multirow{2}{*}{\makebox[\GenTimeWidth]{\textbf{Time}}} & \multirow{2}{*}{\makebox[\GenCostWidth]{\textbf{Cost}}} & \multirow{2}{*}{\makebox[\GenCompletionWidth]{\textbf{Compl.}}} & \multicolumn{5}{c}{\textbf{Per-Level Pass}} & \multirow{2}{*}{\makebox[\GenTokensWidth]{\textbf{\makecell{Actor\\Tok.}}}} & \multirow{2}{*}{\makebox[\GenTimeWidth]{\textbf{Time}}} & \multirow{2}{*}{\makebox[\GenCostWidth]{\textbf{Cost}}} \\
\cmidrule(l{2pt}r{2pt}){3-7}\cmidrule(l{2pt}r{2pt}){12-16}
 &  & \makebox[\GenPassWidth]{\textbf{L1}} & \makebox[\GenPassWidth]{\textbf{L2}} & \makebox[\GenPassWidth]{\textbf{L3}} & \makebox[\GenPassWidth]{\textbf{L4}} & \makebox[\GenPassWidth]{\textbf{L5}} &  &  &  &  & \makebox[\GenPassWidth]{\textbf{L1}} & \makebox[\GenPassWidth]{\textbf{L2}} & \makebox[\GenPassWidth]{\textbf{L3}} & \makebox[\GenPassWidth]{\textbf{L4}} & \makebox[\GenPassWidth]{\textbf{L5}} &  &  &  \\
\midrule
\modellogo{gemini}Gemini 3.6 Flash
& 21.00\% & 50.00\% & 27.50\% & 13.33\% & 9.17\% & 5.00\%
& 389.18M & 28.40h & \$198.62
& 23.50\% & 48.33\% & 31.67\% & 19.17\% & 10.83\% & 7.50\%
& 602.72M & 51.70h & \$328.79 \\
\modellogo{gemini}Gemini 3.7 Flash
& 25.33\% & 55.00\% & 33.33\% & 19.17\% & 10.83\% & 8.33\%
& 486.25M & 46.10h & \$215.64
& 33.67\% & 62.50\% & 40.83\% & 30.00\% & 19.17\% & 15.83\%
& 901.82M & 102.20h & \$394.84 \\
\modellogo{gemini}Gemini 3.8 Flash
& 21.17\% & 51.67\% & 28.33\% & 14.17\% & 8.33\% & 3.33\%
& 795.78M & 62.90h & \$269.20
& 30.67\% & 61.67\% & 39.17\% & 25.00\% & 15.83\% & 11.67\%
& 1741.21M & 142.50h & \$569.26 \\
\addlinespace[2pt]
\bottomrule
\end{tabular}}
\end{table}

\end{document}